%% file: main.tex
\pdfoutput=1
\documentclass[11pt]{article}
\usepackage[utf8]{inputenc}
\usepackage{fullpage}
\input packages.tex

\input macros.tex

\newcommand{\OURS}{\textsc{HAP}\xspace}
\newcommand{\NIOURS}{\textsc{HAP+Implant}\xspace}

\begin{document}
%%%%%%%%% TITLE
\title{Hessian-Aware Pruning and Optimal Neural Implant}

\author{
\large
Shixing Yu$^{*,1}$\thanks{\noindent$^{*}$Equal contribution. \newline$^\dagger$Correspondence to: Amir Gholami: amirgh@berkeley.edu},
Zhewei Yao$^{*,2}$
Amir Gholami$^{*,\dagger,2}$,
Zhen Dong$^{*,2}$,\\
Sehoon Kim$^{2}$,
Michael W. Mahoney$^{2}$,
Kurt Keutzer$^{2}$\\
$^{1}$Peking University,
$^{2}$University of California, Berkeley \\ 
{\tt\scriptsize yushixing@pku.edu.cn,\  \{zheweiy, amirgh, zhendong, sehoonkim, mahoneymw, keutzer\}@berkeley.edu}
}

\maketitle

\input{_s0_abstract.tex}
\input{_s1_introduction}
\input{_s2_related_work}

\input{_s3_method}
\input{_s4_result}

\input{_s5_conclusion}

\section*{Acknowledgments}
The UC Berkeley team also acknowledges gracious support from Samsung (in particular Joseph Hassoun), Intel corporation, Intel VLAB team, Google TRC team, and Google Brain (in particular Prof. David Patterson, Dr. Ed Chi, and Jing Li).
Amir Gholami was supported through through funding from Samsung SAIT.
Our conclusions do not necessarily reflect the position or the policy of our sponsors, and no official endorsement should be inferred.

\small{
\bibliographystyle{ieee_fullname}
\bibliography{ref}
}

\clearpage
\onecolumn
\normalsize
\appendix
\input{_appendix}

\end{document}

%% file: packages.tex
\usepackage[square,sort&compress,comma,numbers]{natbib}

\usepackage{times}
\usepackage{epsfig}
\usepackage{graphicx}
\usepackage{amsmath}
\usepackage{amssymb}
\usepackage{adjustbox}

\usepackage{microtype}
\usepackage{graphicx}
\usepackage{subfigure}
\usepackage{booktabs} % for professional tables

\usepackage[colorlinks=true,
citecolor=blue,
filecolor=black,
linkcolor=blue,
urlcolor=blue]{hyperref}
\input{math_commands.tex}

\usepackage{url}
\usepackage{multirow}
\usepackage{paralist}

\usepackage{booktabs} % for professional tables
\usepackage{multicol}

\usepackage[ruled,noend]{algorithm2e}

\SetCommentSty{mycommfont}

\usepackage{here}
\usepackage{caption}
\usepackage{amsmath,amssymb,amsfonts,amsbsy,amsfonts,latexsym}
\usepackage{amsthm}
\usepackage{multirow}
\usepackage{makecell}
\usepackage[labelfont=bf,textfont=it,belowskip=0pt,aboveskip=5pt,tableposition=top]{caption}
\usepackage{xcolor}
\usepackage{colortbl}

\definecolor{colorA}{RGB}{189,201,225}
\definecolor{colorB}{RGB}{103,169,207}
\definecolor{colorC}{RGB}{ 28,144,153}
\definecolor{colorD}{RGB}{  1,108, 89}

\newcolumntype{R}{>{\columncolor{gray!40}}r}
\newcolumntype{L}{>{\columncolor{gray!40}}l}
\newcolumntype{C}{>{\columncolor{gray!40}}c}

\usepackage{tabularx,colortbl,xcolor}
\usepackage{multirow}
\usepackage[normalem]{ulem}
\useunder{\uline}{\ul}{}

\usepackage{enumitem}

\usepackage{xparse}% http://ctan.org/pkg/xparse

\graphicspath{{fig}}
\DeclareGraphicsExtensions{.pdf,.png}

\SetKwInput{KwInput}{Input}

%% file: math_commands.tex
\usepackage{amsmath,amsfonts,bm}

\usepackage{xcolor}
\usepackage{colortbl}

\def\secref#1{section~\ref{#1}}
\def\eqref#1{equation~\ref{#1}}
\def\1{\bm{1}}

\DeclareMathAlphabet{\mathsfit}{\encodingdefault}{\sfdefault}{m}{sl}
\SetMathAlphabet{\mathsfit}{bold}{\encodingdefault}{\sfdefault}{bx}{n}

\newcommand{\E}{\mathbb{E}}

\newcommand{\R}{\mathbb{R}}

%% file: macros.tex
\NewDocumentCommand{\var}{O{s} m O{}}{%
  \ensuremath{#1_{#2}^{#3}}% add \vphantom{<bizarre sup>}
}
\usepackage{siunitx}

\definecolor{light-gray}{gray}{0.80}

\renewcommand\paragraph{\subsubsection*}

\newcommand\eref{Eq.~\ref}
\newcommand\fref{Figure~\ref}
\newcommand\tref{Table~\ref}

\newcommand\ha{ \rowcolor{orange!0}}

\newcommand\hc{ \rowcolor{orange!40}}

\def\0{{\bf 0}}

\def\R{{\mathbb R}}
\newcommand{\Loss}{\mathcal{L}}

\usepackage{amsthm}

\newtheorem{mytheorem}{Theorem}
\newtheorem{assumption}[mytheorem]{Assumption}
\newtheorem{lemma}[mytheorem]{Lemma}
\newtheorem{remk}[mytheorem]{Remark}

%% file: _s0_abstract.tex
\begin{abstract}
Pruning is an effective method to reduce the  memory footprint  and FLOPs  associated with 
neural network models. However, existing structured-pruning methods often result in significant 
accuracy degradation for moderate pruning levels. 
To address this problem, we introduce a new Hessian Aware Pruning (\OURS) method coupled 
with a Neural Implant approach that uses second-order sensitivity as a metric for 
structured pruning. The basic idea is to prune insensitive components and to use a
Neural Implant for moderately sensitive components, instead of completely pruning them.
For the latter approach, the moderately sensitive components are replaced with  with a low 
rank implant that is smaller and less computationally expensive than the original 
component. We use the relative Hessian trace to measure sensitivity, as opposed to the
magnitude based sensitivity metric commonly used in the literature. 
We test \OURS for both computer vision tasks and natural language tasks, and we achieve new state-of-the-art 
results. Specifically, \OURS achieves less than $0.1\%$/$0.5\%$ degradation on PreResNet29/ResNet50
(CIFAR-10/ImageNet) with more than 70\%/50\% of parameters pruned. 
Meanwhile, \OURS also achieves significantly better performance (up to 0.8\% with 60\% of parameters pruned) as compared to gradient based method for head pruning on transformer-based models.
The framework has been open sourced and available online~\cite{HAP}.
\end{abstract}

%% file: _s1_introduction.tex
\section{Introduction}
\label{sec:intro}

There has been a significant increase in the computational
resources required for Neural Network (NN) training and inference.
This is in part due to larger input sizes (e.g., higher image resolution) as well
as larger NN models requiring more computation with a significantly larger memory footprint.
The slowing down of Moore's law, along with challenges associated with increasing
memory bandwidth, has made it difficult
to deploy these models in practice.
Often, the inference time and associated power consumption is orders
of magnitude higher than acceptable ranges.
This has become a challenge for many applications, e.g., health care and personalized medicine, which
have restrictions on uploading data to cloud servers, and which have to rely
on local servers with limited resources. Other applications include
inference on edge devices such as mobile processors, security cameras, and
intelligent traffic control systems, all of which require real-time inference.
Importantly, these problems are not limited to edge devices, and state-of-the-art models for applications such
as speech recognition, natural language processing, and recommendation systems often cannot be efficiently performed
even on high-end servers.

A promising approach to address this is pruning.
~\cite{luo2017thinet,liu2018rethinking,dong2017learning,zeng2018mlprune,lecun1990optimal,han2015learning, molchanov2016pruning, li2016pruning, mao2017exploring,yang2017designing,wang2020differentiable,tung2018clip,guerra2020automatic,hacene2018quantized,zhuang2018discrimination,liu2019metapruning,kwon2020structured,wang2019eigendamage,zhu2017prune,guo2020dmcp, li2020eagleeye},
However, an important challenge is determining which parameters are insensitive to the pruning process. A brute-force
method is not feasible since one has to test each parameter in the network separately and measure
its sensitivity. The seminal work of~\cite{lecun1990optimal} proposed Optimal Brain Damage (OBD), a second-order based method
to determine insensitive parameters. However, this approach requires pruning the parameters one at a time, which is
time-consuming. To address this problem, we propose a simple, yet effective, modification of OBD by using the Hessian trace
to prune a group of parameters along with a low rank Neural Implant.
In more detail, our contributions are as follows:
\begin{itemize}[noitemsep,topsep=0pt,parsep=0pt,partopsep=0pt,leftmargin=*]
    \item We propose \OURS, a Hessian Aware Pruning method that uses a fast second-order metric to find insensitive
    parameters in a NN model. In particular, we use the average Hessian trace to weight the magnitude of the parameters in the NN.
    Parameters with large second-order sensitivity remain unpruned, and those with relatively small sensitivity are pruned.
    In contrast to OBD~\cite{lecun1990optimal}, \OURS 
    finds groups of insensitive parameters, which is faster than pruning a single parameter at a time.
    Details of the \OURS method are discussed in Section~\ref{sec:method}.
    
    \item We propose a novel Neural Implant (denoted by \NIOURS) technique to alleviate accuracy degradation. In this approach, we replace moderately sensitive model
    components with a low rank implant. The model along with the implant is then
    fine-tuned. We find that this approach helps boost the accuracy.
    % Specifically, instead of pruning the entire insensitive kernels, we replace them with $1\times1$ convolutions.
    % These implanted pointwise convolutions are then trained, and this helps boost the accuracy.
    For details, see Section~\ref{subsec:decompose}.
    
    \item We perform detailed empirical testing and show that \OURS achieves 94.3\% accuracy ($<0.1\%$ degradation) on PreResNet29 (CIFAR-10), with only 31\% parameters left (\fref{fig:cifar10_pareto}).
    In comparison to EigenDamage, a recent second-order pruning method, we achieve up to 1.2\% higher accuracy with fewer parameters and FLOPs (\fref{fig:cifar10_pareto}).
    Moreover, for ResNet50, \OURS achieves 75.1\% top-1 accuracy (0.5\% degradation) on ImageNet, with only half of the parameters left (\tref{tab:dhap}).
    In comparison to prior state-of-the-art HRank~\cite{lin2020hrank}, \OURS achieves up to 2\% higher accuracy with fewer parameters and FLOPs (\tref{tab:dhap}).
    For head pruning of RoBERTa on MRPC/QNLI, \OURS achieves up to 0.82\%/0.89\% higher accuracy than the gradient based method~\cite{michel2019sixteen} (\tref{tab:roberta_result}).
    \item We perform detailed ablation experiments to illustrate the efficacy of the second-order sensitivity metric. In particular, we compare the second-order sensitivity with
    a random method, and a reverse-order in which the opposite order sensitivity order given by \OURS is used.
 In all cases, \OURS achieves higher accuracy (\tref{tab:ablation}).
\end{itemize}

%%%%%%%%%%%%%%%%%%%%%%%%%%%%%%%%%%%%%%%%%%%%%%%%%%%%%%%%%%%%%%%%%%%%%%%%%%%%%%%%%%%
\begin{figure}[t]
\centering
\includegraphics[width=1.0\textwidth]{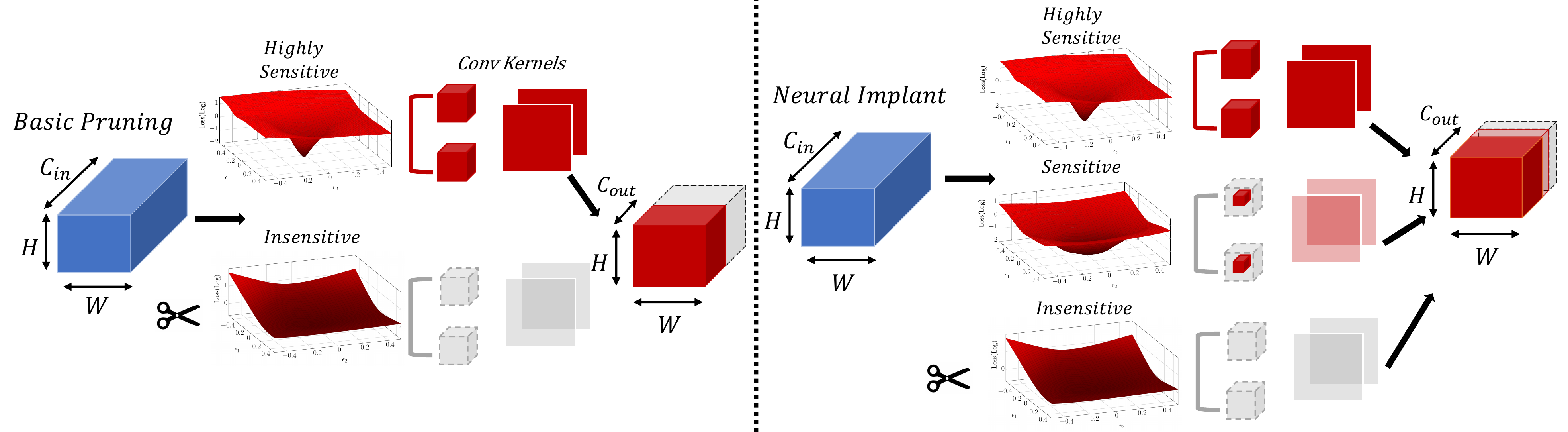}
\caption{
(Left) The \OURS method is a structured pruning method that prunes channels based on their second-order sensitivity, which measured flatness/sharpness of loss landscape.
Channels are sorted based on this metric, and only insensitive channels are pruned.
(Right) Similar to other
structured-pruning methods, \OURS  at large pruning ratios results in accuracy degradation. This because one has to inevitably prune moderately sensitive channels
at high pruning ratios, which may contain individual neurons that are very sensitive.
The removal of the entire channel, along with the sensitive neurons results in
accuracy degradation.
To address this, we propose \NIOURS method, where such channels are
replaced with a light-weight, low-rank \textit{Neural Implant}.
}
\label{fig:channel_condensing}
\end{figure}
%%%%%%%%%%%%%%%%%%%%%%%%%%%%%%%%%%%%%%%%%%%%%%%%%%%%%%%%%%%%%%%%%%%%%%%%%%%%%%%%%%%

%% file: _s2_related_work.tex
\section{Related work}
\label{sec:related_work}

Several different approaches have been proposed to make NN models more efficient by making
them more compact, faster, and more energy efficient. These efforts could be generally
categorized as follows:
(i) efficient NN design~\cite{iandola2016squeezenet,howard2017mobilenets, sandler2018mobilenetv2, howard2019searching,zhang2018shufflenet,ma2018shufflenet};
(ii) hardware-aware NN design~\cite{gholami2018squeezenext,wu2019fbnet,cai2019once,lyna2020fast, yang2018netadapt, tan2019mnasnet};
(iii) quantization~\cite{wu2016quantized, jacob2018quantization, hubara2017quantized,krishnamoorthi2018quantizing, dong2019hawq, dong2019hawqv2};
(iv) distillation~\cite{hinton2015distilling, mishra2017apprentice, yin2020dreaming, polino2018model};
and (v) pruning.
% Pruning, especially structured-pruning, can be an effective way to accelerate inference.

% \subsection{Structured Pruning}
Here we briefly discuss the related work on pruning, which can be broadly
categorized into: unstructured pruning~\cite{dong2017learning, lee2018snip, xiao2019autoprune, park2020lookahead}; and structured pruning~\cite{luo2017thinet, he2018amc, yu2018nisp, lin2018accelerating, huang2018data, zhao2019variational}.
Unstructured pruning prunes out neurons without any structure.
However, this leads to sparse matrix operations which are hard to accelerate 
and are typically memory-bounded
~\cite{buluc2008challenges,gale2019state}.
This can be addressed with structured pruning, where an entire matrix operation (e.g., an output channel) is removed. However, the challenge here is that high degrees of structured
pruning often leads to significant accuracy degradation.

In both approaches, the key question is to find which parameters to prune.
A simple and popular approach is magnitude-based pruning. 
In this approach, the magnitude of parameters is used as the pruning metric.
The assumption here is that small parameters are not important and can be removed.
A variant of this approach was used in~\cite{liu2017learning}, where 
the scaling factor of the batch normalization layer is used as the sensitivity metric.
In particular, channels with smaller scaling factors (or output values) are considered less important and got pruned.
Another variation is proposed by~\cite{li2016pruning}, where channel-wise summation over weights is used as the metric.
Other methods have been proposed as alternative sensitivity metrics. 
For instance, \cite{lin2020hrank} uses channel rank as sensitivity  metric; \cite{he2017channel} uses a LASSO regression based channel selection criteria; and~\cite{he2019filter} uses the geometric median of the convolutional filters.
An important problem with magnitude-based pruning methods is that parameters with small magnitudes can actually be quite sensitive.
It is easy to see this through a second-order Taylor series expansion, where the perturbation is dependent on not just the weight magnitude but
also the Hessian~\cite{lecun1990optimal}. In particular, small parameters with large Hessian could in fact be very sensitive, as opposed to large parameters with small Hessian (here, we are using small/large Hessian loosely; the exact metric to measure is given by the second-order perturbation in~\eref{eq:objective}).
For this reason, OBD~\cite{lecun1990optimal} proposes to use the Hessian diagonal as the sensitivity metric.
The follow up work of Optimal Brain Surgeon (OBS)~\cite{hassibi1993optimal,hassibi1993second} used a similar method, but considered off-diagonal Hessian components,
and showed a correlation with inverse Hessian.
One important challenge with these methods is that pruning has to be performed one parameter at a time.
The recent work of~\cite{dong2017learning} extends this to layer-wise pruning in order to reduce the cost of computing Hessian information for one parameter at a time.
However, this method can result in unstructured pruning.
Another second-order pruning method is EigenDamage~\cite{wang2019eigendamage}, where the Gauss-Newton operator is used instead of Hessian.
In particular, the authors use Kronecker products to approximate the GN operator.
Our findings below show that using the average Hessian trace method significantly outperforms EigenDamage.
We also find that it is very helpful to replace moderately sensitive layers
with a low rank Neural Implant, instead of completely pruning them, as discussed~next.

%% file: _s3_method.tex
\section{Methodology}
\label{sec:method}
Here, we focus on supervised learning tasks, where the nominal goal is to minimize
the empirical risk by solving the following optimization problem:

\begin{equation}
\small
    L(w) = \frac{1}{N} \sum\nolimits_{i=1}^N l(x_i, y_i, w),
    \label{eq:erm}
\end{equation}
where $w\in \R^n$ is the trainable model parameters, $l(x_i, y_i, w)$ is the loss for the input datum $x_i$, where $y_i$ is the corresponding label, and $N$ is the  training set cardinality. 
For pruning, we assume that the model is already trained and converged
to a local minima which satisfies the first and second-order
optimality conditions (that is, the gradient $\nabla_{w}L(w)=0$, and the Hessian is Positive Semi-Definite (PSD), $\nabla^2_{w}L(w) \succcurlyeq 0$).
The problem statement is to prune (remove) as many parameters as possible to reduce
the model size and FLOPs to a target threshold with minimal accuracy degradation.

We first start with a general description of the problem and then derive our method.
Let $\Delta w \in \R^n$ denote the pruning perturbation 
such that the corresponding weights become zero (that is $w+\Delta w = 0$).
We denote the corresponding change of loss as $\Delta L$:
\begin{equation}
\small
\label{eq:taylor}
    \Delta L = L(w+\Delta w) - L(w) = g ^T \Delta w + \frac{1}{2} \Delta w ^T H \Delta w + O(||\Delta w||^3). 
\end{equation}
where the second equation comes from Taylor series expansion. Here $g\in \R^{n}$ denotes the gradient of loss function $L$ w.r.t. weights $w$, and $H \in \R^{n\times n}$ is the corresponding Hessian operator (i.e. second-order
derivative). 
For a pretrained neural network that has already converged to a local minimum, we have $g = 0$, and the Hessian is a PSD matrix.
As in prior work~\cite{hassibi1993optimal}, we assume higher-order terms, e.g., $O(||\Delta w||^3)$, in~\eref{eq:taylor} can be ignored.

The pruning problem is to find the set of weights that result in minimum perturbation
to the loss ($\Delta L$). This leads to the following constrained optimization problem:
\begin{equation}
\scriptsize
\label{eq:objective}
\begin{split}
    \min_{\Delta w} \frac{1}{2}\Delta w ^T H \Delta w 
    &=\frac{1}{2}\begin{pmatrix} \Delta w_{p}\\ \Delta w_{l} \end{pmatrix}^T 
    \begin{pmatrix} H_{p,p} & H_{p,l} \\ H_{l,p} & H_{l,l} \end{pmatrix}
    \begin{pmatrix} \Delta w_{p}\\ \Delta w_{l} \end{pmatrix},\\
    &\text{s.t.} \quad \Delta w_{p} + w_p = 0.
\end{split}
\end{equation}
Here, we denote the channels that are pruned with $p$ as the subscript (e.g. $w_p\in \R^{p}$), and denote the remaining parameters with $l$ as the subscript (e.g. $w_l\in \R^{n-p}$). 
Similarly, we use $\Delta w_{p}$ and $\Delta w_{l}$ to denote the corresponding perturbations.
Note that $\Delta w_p=-w_p$ since $p$-channels are pruned.
Moreover, $H_{l,p}$ denotes the cross Hessian w.r.t. $l$-channels and $p$-channels (and 
similarly $H_{p,p}$ and $H_{l,l}$ are Hessian w.r.t. pruned and unpruned parameters).
Using Lagrangian method, we can finally get (see~\secref{sec:lagrangian} for more details):
\begin{equation}
    \label{eqn:sensitivity_1}
\small
    \frac{1}{2} \Delta w ^T H \Delta w = \frac{1}{2} w_{p}^T (H_{p, p} - H_{p, l}  H_{l, l}^{-1} H_{l, p}) w_{p}.
\end{equation}
% \normalsize
\eref{eqn:sensitivity_1} gives us the perturbation to the loss when a
set of parameters $w_p$ is removed.
It should be noted that OBS~\cite{hassibi1993optimal} and L-OBS~\cite{dong2017learning}, where OBS is applied for each layer under the assumption of cross-layer independence, is a degenerate case of~\eref{eqn:sensitivity_1} for the
special case of  $w_p \in \R^1$. Next we discuss how this general formulation
can be simplified.

%-----------------------------------------------
% \vspace{-2mm}
\subsection{Hessian-aware Pruning}
\label{subsec:hap}
% \vspace{-2mm}

There are three major disadvantages with OBS.
First, computing~\eref{eqn:sensitivity_1} requires computing (implicitly) information from the inverse Hessian, $H_{l,l}^{-1}$.
This can be costly, both in terms of computations and memory (even when using matrix-free randomized methods).
The work of L-OBS~\cite{dong2017learning} attempted to address this challenge by ignoring cross-layer dependencies, but it still requires computing block-diagonal inverse Hessian information, which can be costly. % (just not Hessian information for the entire network).
Second, in both OBS and L-OBS, one has to measure this perturbation for all the parameters separately, and then prune those parameters that result in the smallest perturbation. 
This can have a high computational cost, especially for deep models with many parameters. 
Third, this pruning method results in unstructured pruning, which is difficult to accelerate
with current hardware architectures. 
In the OBD~\cite{lecun1990optimal} method the first problem does not exist as the
the Hessian is approximated as a diagonal operator, without the need to compute
inverse Hessian:
\begin{equation}
\small
\begin{split}
        \frac{1}{2} \Delta w ^T H \Delta w 
    % & = \frac{1}{2} w_{p}^T (H_{p, p} - H_{p, l}  H_{l, l}^{-1} H_{l, p}) w_{p}.\\
    &\approx \frac{1}{2} w_{p}^T Diag(H_{p, p})  w_{p}.
    \label{eqn:sensitivity_2}
\end{split}
\end{equation}
Here $Diag(H_{p,p})$ denotes the diagonal elements of $H_{p,p}$. 
However, the second and third of these disadvantages still remain with OBD.

%%%%%%%%%%%%%%%%%%%%%%%%%%%%%%%%%%%%%%%%%%%%%%%%%%%%%%%%%%%%%%%%%%%%%%%%%%%%%%%%%%%
\begin{figure*}[t]
\centering
\includegraphics[trim=33 27 33 33, clip, width=.32\textwidth]{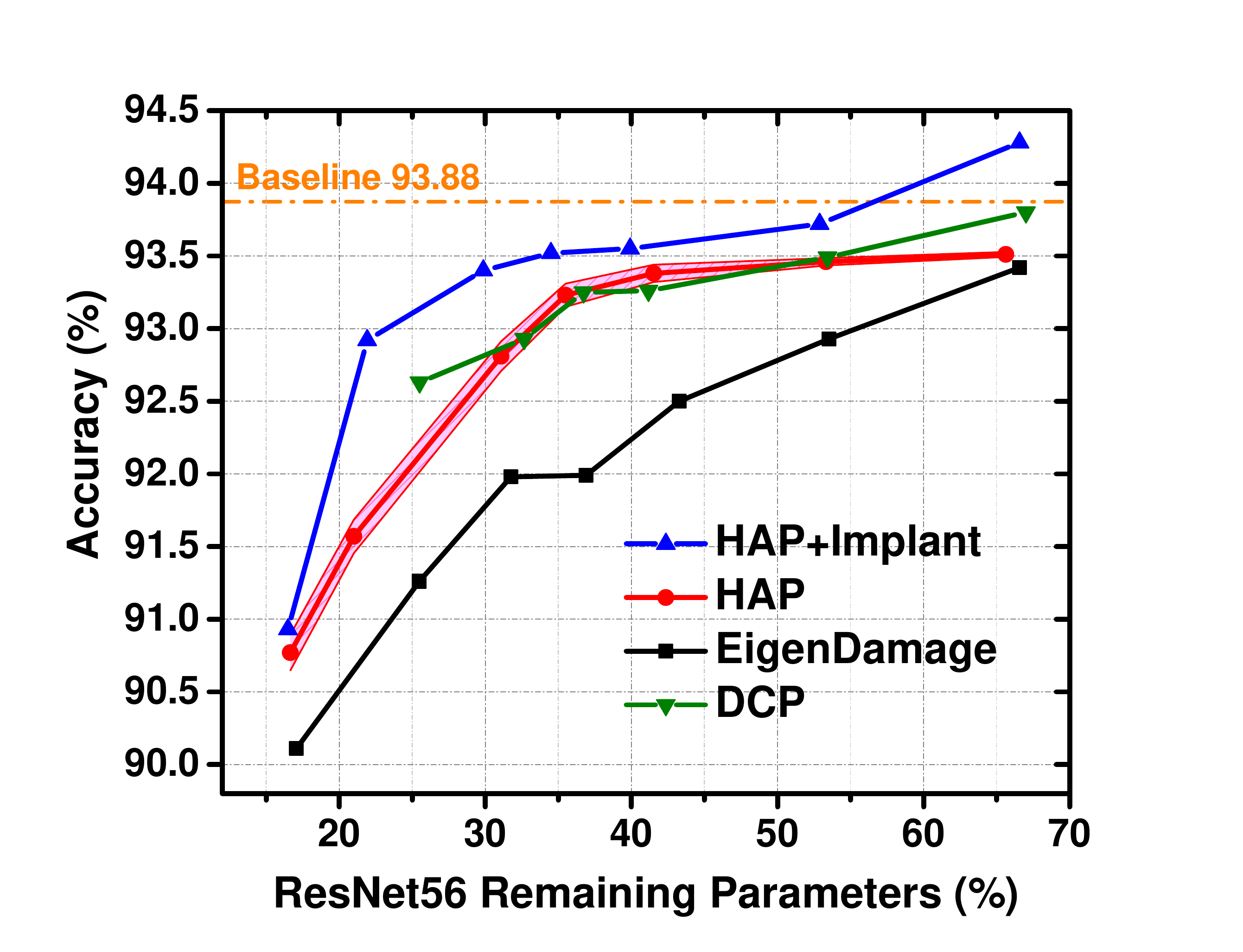}
\includegraphics[trim=33 27 33 33, clip, width=.32\textwidth]{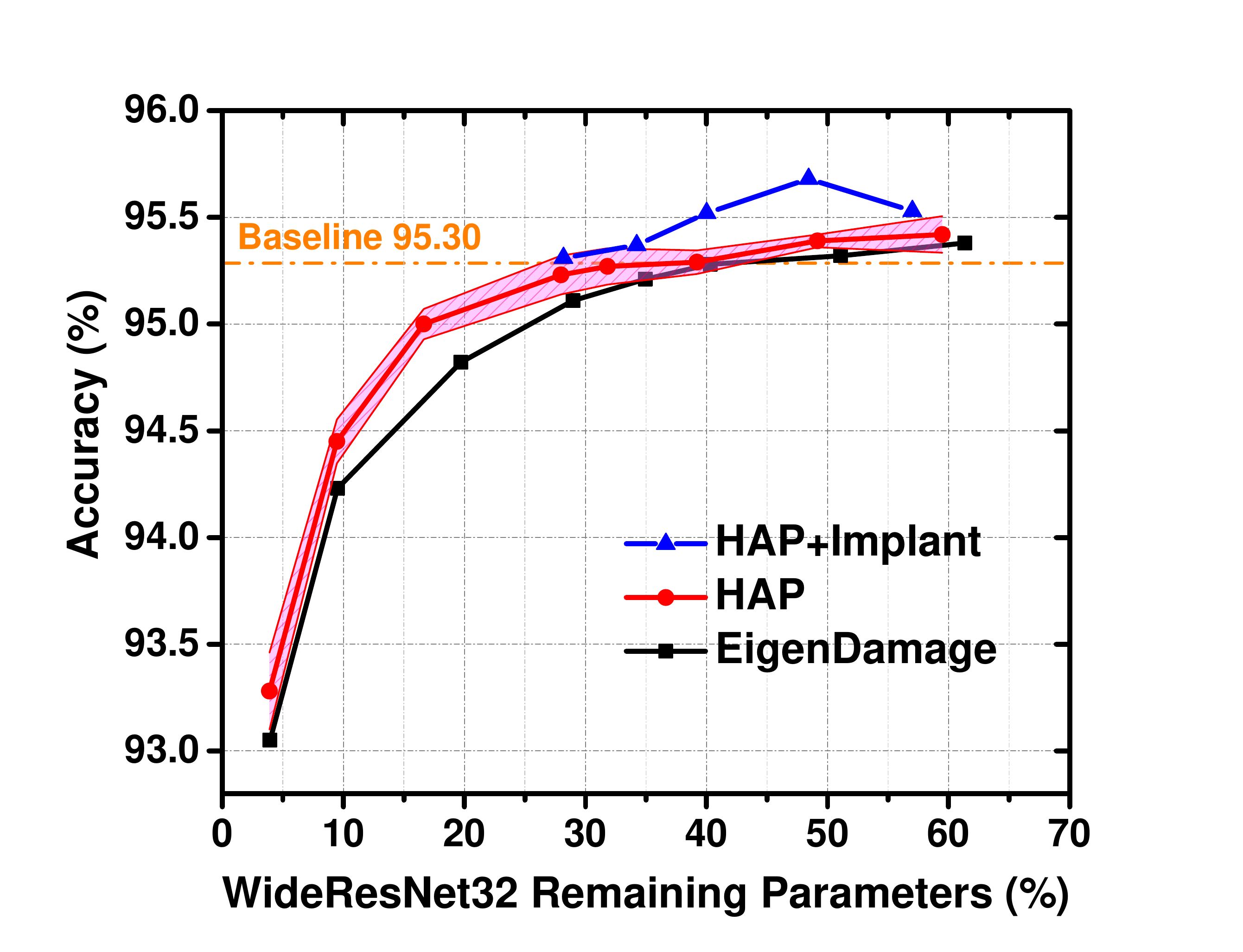}
\includegraphics[trim=33 27 33 33, clip, width=.32\textwidth]{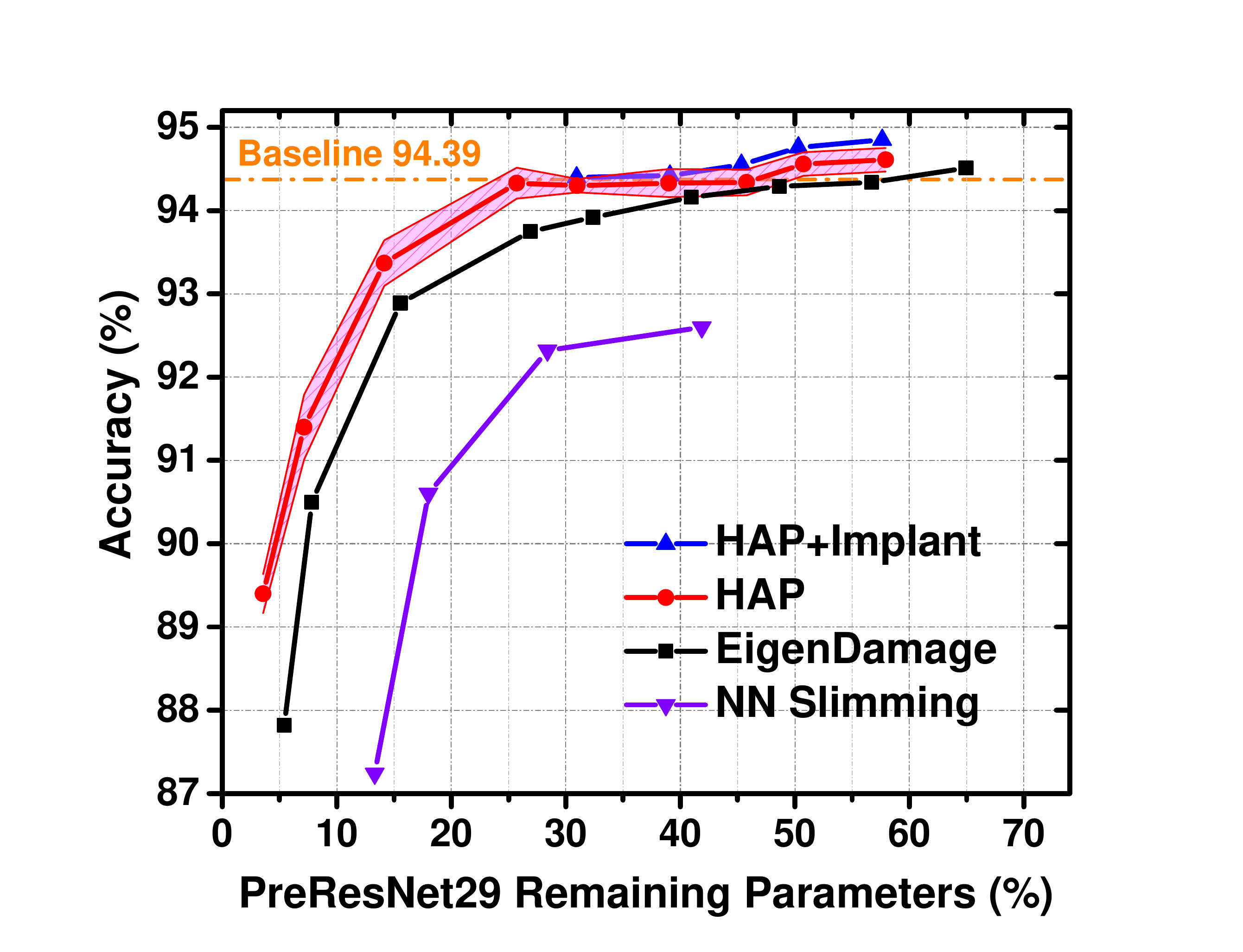}\\
\includegraphics[trim=33 27 33 33, clip, width=.32\textwidth]{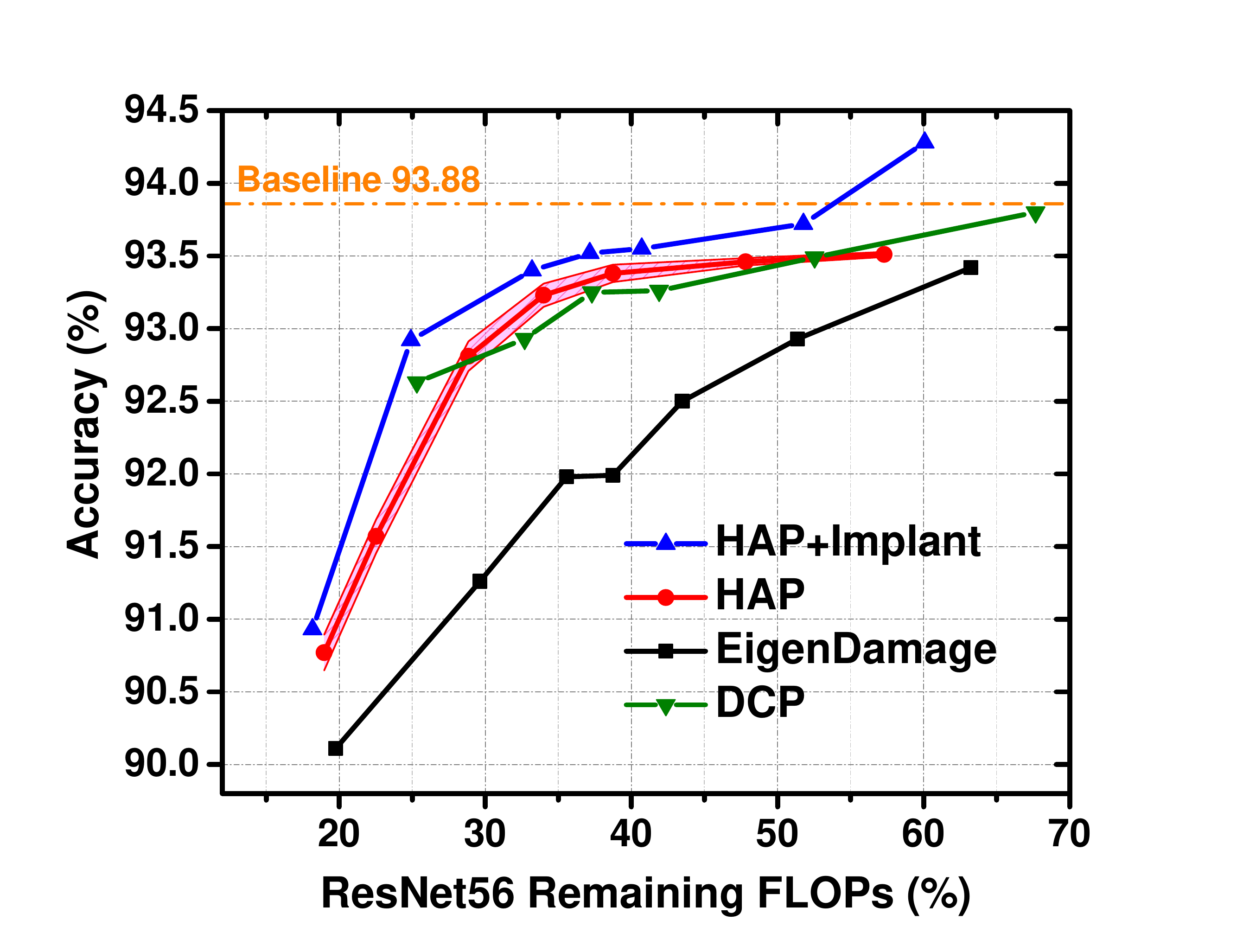}
\includegraphics[trim=33 27 33 33, clip, width=.32\textwidth]{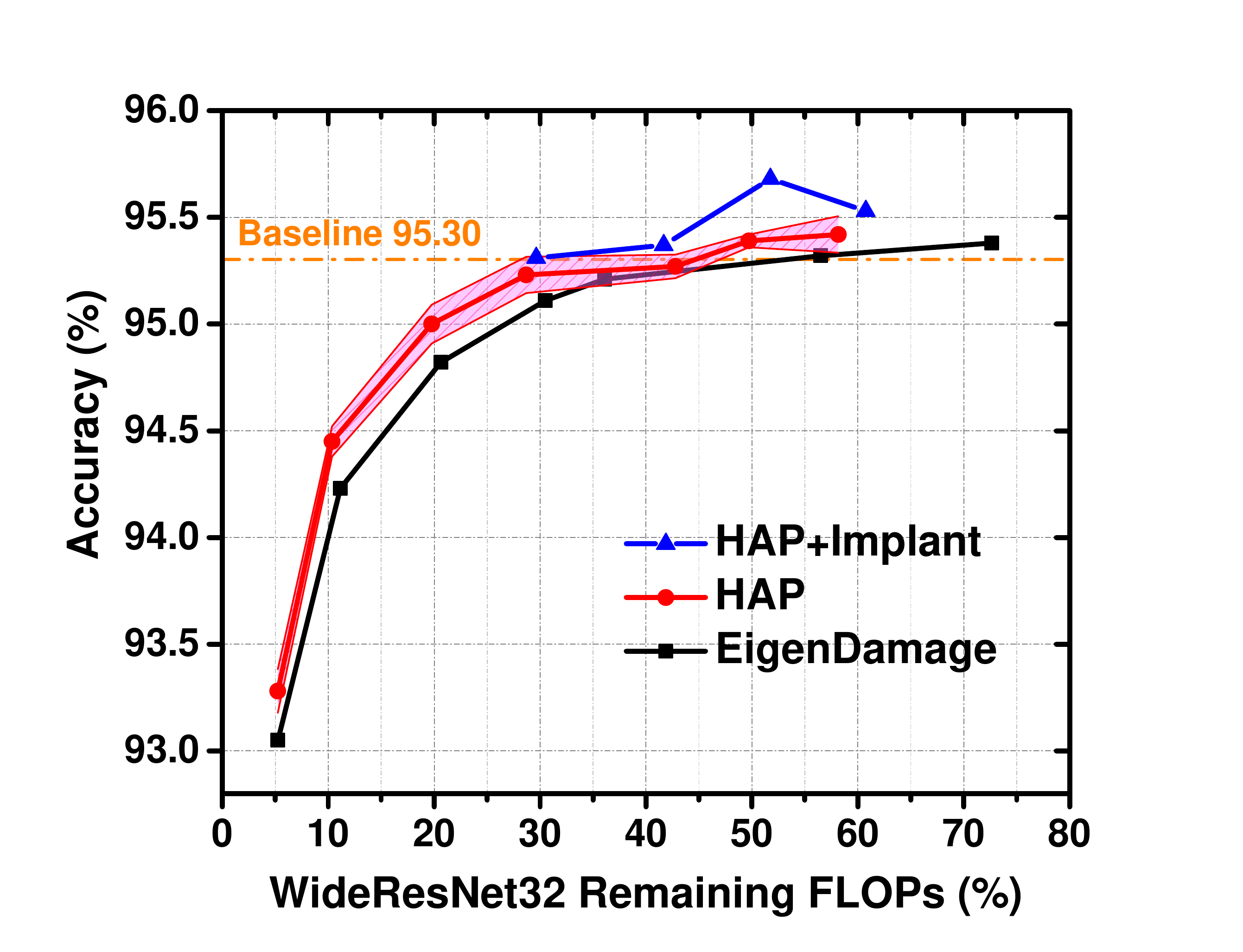}
\includegraphics[trim=33 27 33 33, clip, width=.32\textwidth]{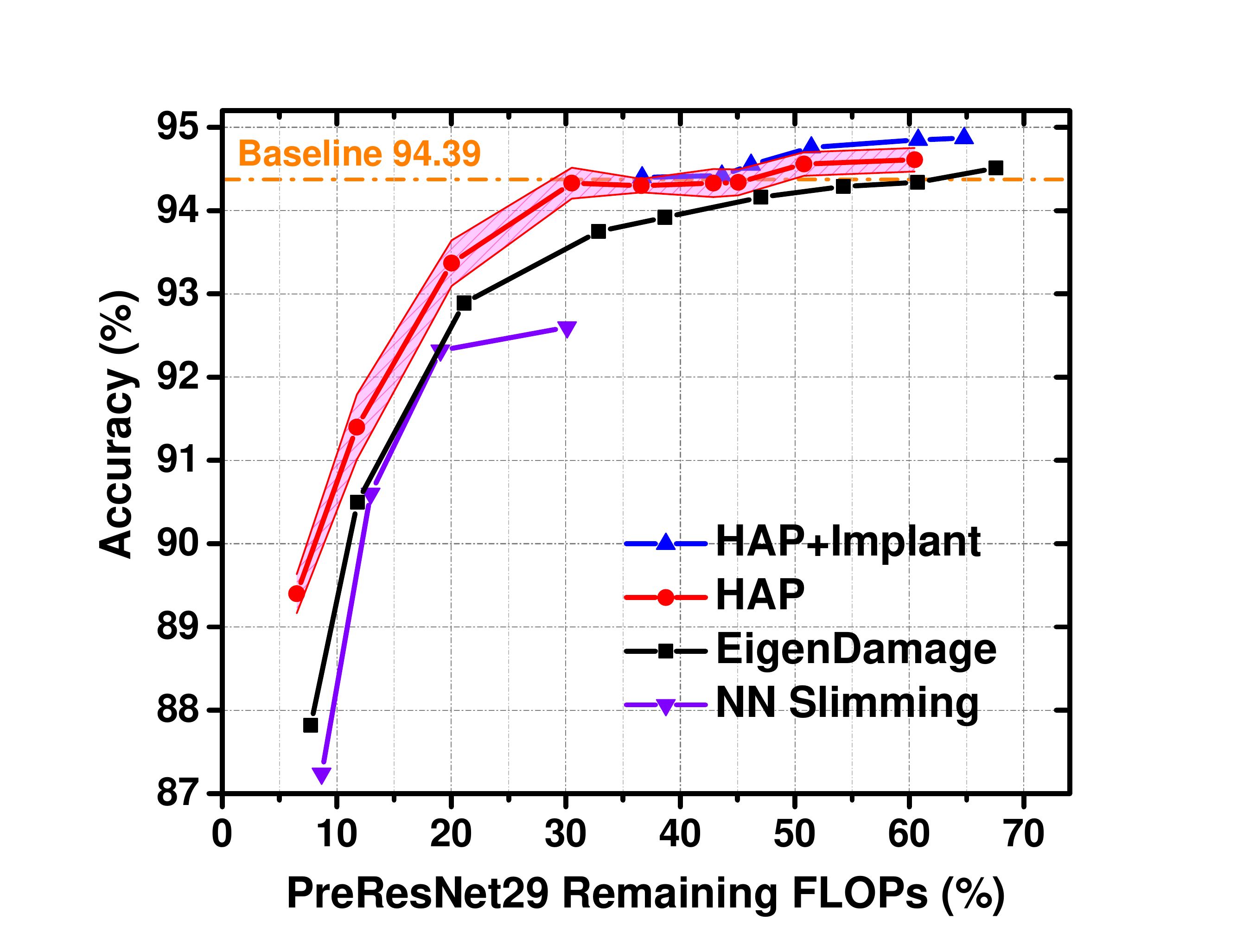}
% \vspace{-2mm}
\caption{
Comparison of accuracy with different pruning ratios among \NIOURS, \OURS, NN Slimming, EigenDamage, and DCP, on the CIFAR-10 dataset, for ResNet56, WideResNet32, and PreResNet29.
(Top) Remaining parameters in the network after pruning is used for x-axis.
(Bottom) Remaining FLOPs in the network after pruning is used for x-axis.
\OURS consistently outperforms EigenDamage and NN Slimming, and \NIOURS boosts performance for moderate pruning ratios and surpasses DCP.
% \amir{@Shixing/Zhen add the error bars}
}
\label{fig:cifar10_pareto}
% \vspace{-4mm}
\end{figure*}
%%%%%%%%%%%%%%%%%%%%%%%%%%%%%%%%%%%%%%%%%%%%%%%%%%%%%%%%%%%%%%%%%%%%%%%%%%%%%%%%%%%

To address the second and third of these disadvantages,
we propose to group the parameters and to compute
the corresponding perturbation when that group is pruned, rather than computing the perturbation for every single parameter separately. 
Note that this can also address the third disadvantage, since pruning a group of parameters (for example parameters in a convolution channel) results in structured pruning.
This can be achieved by considering the Hessian as a block diagonal operator, and then approximating each block with a diagonal operator, with Hessian trace as the diagonal entries. 
In particular, we use the following approximation:
\begin{equation}
\label{eq:trace_approximation}
    \small
    \begin{split}
    \frac{1}{2} \Delta w ^T H \Delta w & = \frac{1}{2} w_{p}^T [H^{-1}]_{p, p}^{-1} w_{p} 
    \approx \frac{1}{2} w_{p}^T \frac{Trace(H_{p,p})}{p} w_{p} 
    = \frac{Trace(H_{p,p})}{2p}\|w_p\|_2^2,
    \end{split}
\end{equation}
where $Trace(H_{p,p})$ denotes the trace of the block diagonal Hessian (the corresponding Hessian block for pruned parameters $H_{p,p}$).
The Hessian can be computed very efficiently with randomized numerical linear algebra methods, in particular Hutchinson's 
method~\cite{bai1996some,avron2011randomized,yao2019pyhessian,yao2020adahessian}.
Importantly, this approach requires computing only the application of the Hessian to a random input vector.
This has the same cost as back-propagating the gradient~\cite{yao2019pyhessian,yao2020adahessian}. 
(Empirically, in our experiments corresponding to ResNet50 on ImageNet, the longest time for computing this trace was three minutes.)
A similar approach was proposed by~\cite{dong2019hawqv2} in the context of quantization.

In more detail, \OURS performs structured pruning by grouping the parameters and approximating the corresponding Hessian as a diagonal operator, with the average Hessian trace of that group as its entries.
For a convolutional network, this group can be an output channel.
We found that this simple modification results in a fast and efficient
pruning method that when combined with the Neural Implant approach
exceeds state-of-the-art. This is discussed~next.

%-------------------------------------------------------------------------
% \vspace{-2mm}
\subsection{Hessian-aware Neural Implant}
\label{subsec:decompose}
% \vspace{-2mm}

In \OURS, we sort the channels from most sensitive to least sensitive (based on~\eref{eq:trace_approximation}).
For a target model size or FLOPs budget, one has then to prune by starting from insensitive
channels.
This approach works well, as long as all these channels are extremely insensitive. However, in practice, some of the sorted
channels will exhibit some level of sensitivity. Entirely pruning these channels,
and leaving the rest of the sensitive ones unpruned, can result in significant accuracy
degradation. This is one of the major problems with structured pruning methods, as
very few groups of parameters are completely insensitive. When those are pruned,
the remaining groups/set of parameters always include some subset of highly sensitive
neurons that if pruned, would result in high accuracy loss.

Here, we propose an alternative strategy to replace moderately sensitive parameter groups
with a low rank \emph{Neural Implant}.
The basic idea is to prune insensitive layers completely, but detect the moderately sensitive 
layers, and instead of completely removing all of its parameters (which can contain
some sensitive ones as discuss above), replace them with a low rank decomposition.
As an example, a spatial convolution could be replaced with a new point-wise convolution
that has smaller parameters and flops. One could also consider other types of
low rank decomposition (e.g. CP/Tucker decomposition, depth-wise/separable convolution,etc). However, for simplicity
we only use a pointwise convolution implant in this paper.

After the implant, the model is fine-tuned to recover accuracy.
We denote this approach as \NIOURS, which is schematically illustrated in~\fref{fig:channel_condensing}.
In summary, we use the Hessian metric in~\eref{eq:trace_approximation}, and then we apply a Neural Implant to the most sensitive channels to be pruned. 

We have to emphasize that many prior works have investigated low-rank matrix approximation~\cite{Mah-mat-rev_BOOK}.
However, existing methods for NN pruning replace all or part of the model, irrespective of their sensitivity, whereas in our approach we perform a targeted low-rank approximation, and only replace the sensitive parts of the model, quantified through the Hessian in~\eref{eq:trace_approximation}.
We empirically found that this approach
is quite effective, especially for moderate pruning ratios, as discussed next.

%% file: _s4_result.tex
%%%%%%%%%%%%%%%%%%%%%%%%%%%%%%%%%%%%%%%%%%%%%%%%%%%%%%%%%%%%%%%
% \vspace{-2mm}
\begin{minipage}{\textwidth}
    \begin{minipage}[t]{0.48\textwidth}
    \makeatletter\def\@captype{table}
    \caption{Comparison between \OURS and other pruning methods on CIFAR-10. Here, VGG16 denotes the baseline used in HRank~\cite{lin2020hrank}, and VGG16-\OURS denotes the baseline used by \OURS method. As one can see, \OURS consistently outperforms other pruning methods, even though its pruned models have fewer parameters (Param.) and FLOPs.}
    \begin{center}
    \label{tab:vgg16}
    \begin{adjustbox}{width=1.0\linewidth} 
    \begin{tabular}{lcccccccccccccc} \toprule
    Method & Acc.(\%)    & Param.(\%)  & FLOPs(\%)  \\
    \midrule
    \ha  VGG16                          & 93.96         & 100.0         & 100.0     \\ 
    \hc  VGG16-HAP                      & 93.88         & 100.0         & 100.0     \\
    \midrule    
    \ha  L1\cite{li2016pruning}         & 93.40         & 36.0          & 65.7      \\
    \ha  SSS\cite{huang2018data}        & 93.02         & 26.2          & 58.4      \\
    \ha  VarP\cite{zhao2019variational} & 93.18         & 26.7          & 60.9      \\
    \ha  HRank\cite{lin2020hrank}       & 93.43         & 17.1          & 46.5      \\
    \ha  GAL-0.05\cite{lin2019towards}  & 92.03         & 22.4          & 60.4      \\
    \ha  HRank\cite{lin2020hrank}       & 92.34         & 17.9          & 34.7      \\
    \ha  GAL-0.1\cite{lin2019towards}   & 90.73         & 17.8          & 54.8      \\
    \hc  \OURS                          & \bf{93.66}    & \bf{10.1}     & \bf{29.7}      \\
    \midrule
    \ha  HRank\cite{lin2020hrank}       & 91.23         &  8.0          & 23.5      \\
    \hc  \OURS                          & \bf{93.37}    &  \bf{5.1}     & \bf{20.3}      \\
    \midrule
    \hc  \OURS                          & \bf{91.22}    & \bf{1.6}      &\bf{7.5}   \\
    \bottomrule 
    \end{tabular}
    \end{adjustbox}
    \end{center}

    \end{minipage}\hfill
    \begin{minipage}[t]{0.48\textwidth}
    \makeatletter\def\@captype{table}
    \caption{Comparison of FLOPs and accuracy on CIFAR-10 using ResNet56 for different pruning methods.
    We report the baseline accuracy used in each work, as well as the corresponding final accuracy after pruning.
    For ease of comparison, we also report the accuracy drop (Acc. $\downarrow$) w.r.t. each baseline. As one can see, \OURS and \NIOURS consistently outperform other work reported in the literature.
    }
    \label{tab:dhap}
    \begin{center}
    \begin{adjustbox}{width=1.0\linewidth} 
    \setlength\tabcolsep{2.35pt}
    \begin{tabular}{lcccccccccccccc} \toprule

    Method                              & Base-acc. & Final-acc. &  Acc. $\downarrow$ & FLOPs (\%)  \\
        
    \midrule    
        
    \ha  CP\cite{he2017channel}             & 92.80 & 91.80     &  1.00             & 50.0                  \\
    \ha  AMC\cite{he2018amc}                & 92.80 & 91.90     &  0.90             & 50.0                  \\
    \ha  FPGM\cite{he2019filter}            & 93.59 & 93.26     &  0.33             & 47.4                  \\
    \ha  LFPC\cite{he2020learning}          & 93.59 & 93.24     &  0.35             & 47.1                  \\
    \hc \NIOURS                             & \textbf{93.88}    & \textbf{93.55}    & \textbf{0.33} & \bf{40.7}      \\
    \midrule
    \ha  GAL-0.8\cite{lin2019towards}       & 93.26             & 90.36             & 2.90          & 39.8      \\
    \ha  HRank\cite{lin2020hrank}           & 93.26             & 90.72             & 2.54          & 25.9      \\
    \hc  \OURS                              & 93.88             & 91.57             & 2.31          & \textbf{21.0}      \\
    \hc  \NIOURS                            & \bf{93.88}        & \bf{92.92}        & \bf{0.96}     & 23.9      \\
    \bottomrule 
    \end{tabular}
    \end{adjustbox}
    \end{center}
    \end{minipage}\hfill
\end{minipage}
%%%%%%%%%%%%%%%%%%%%%%%%%%%%%%%%%%%%%%%%%%%%%%%%%%%%%%%%%%%%%%%

\section{Results}
\label{sec:results}
\subsection{Experimental Settings}

\textbf{Computer Vision.} For evaluating the performance of \OURS, we conduct experiments for image classification on CIFAR-10 (ResNet56/WideResNet32/PreResNet29/VGG16) and ImageNet (ResNet50). 
Our main target comparison for HAP (without Implant) is EigenDamage, a recent second-order pruning method. For fair comparison, we use
the same pretrained model used by EigenDamage when available (WideResNet32 on Cifar-10),
and otherwise train the model from scratch (ResNet56, VGG16, and PreResNet29 on Cifar-10).
For all cases, we ensure comparable baseline accuracy, and when not possible, we report
the baseline used by other methods.
For comparison we consider a wide range of pruning ratios, and
consider validation accuracy, FLOPs, and parameter size as the metrics. The goal is to achieve higher accuracy with lower FLOPs/parameter size.

\textbf{Natural Language Understanding.} 
We use RoBERTa-base~\cite{liu2019roberta}, which consists of 12 attention heads for each of 12 Transformer encoder layers, as the baseline model.
It has been explored in~\cite{michel2019sixteen} that not all heads in Transformer architectures are equally important and thus a great portion of them can be removed without degrading the accuracy.

\subsection{HAP Results on CIFAR-10}
\label{sec:hap_cifar10_results}
% \vspace{-2mm}

We first start with evaluating \OURS without Neural Implant, and then
discuss the specific improvement of using Neural Implant.
The results on
CIFAR-10 for different
pruning ratios and various models are presented in~\fref{fig:cifar10_pareto}.
In particular, we report both the validation accuracy versus remaining parameters after 
pruning, as well as validation accuracy versus the FLOPs.
For comparison, we also plot the performance of NN slimming~\cite{liu2017learning}, EigenDamage~\cite{wang2019eigendamage} and DCP~\cite{zhuang2018discrimination} for different pruning ratios. 
For all the points that we compare, HAP achieves higher accuracy than 
EigenDamage, even for cases with fewer parameters/FLOPs.
We generally observe that the difference between \OURS and EigenDamage is more noticeable for higher pruning ratios (i.e., fewer remaining parameter). 
This is expected, since small amounts of pruning does not lead to significant accuracy degradation, while higher pruning ratios are more challenging.
In particular, when the parameter remaining percentage is around 35\% (i.e., 65\% of the parameters are pruned), \OURS achieves 93.2\% accuracy, which is 1.24\% higher than EigenDamage,
with fewer FLOPs (34.0\% versus 38.7\% for EigenDamage).
We observe a similar trend on WideResNet32, where \OURS consistently outperforms
EigenDamage.

We also plot the performance of DCP, which is not a second-order
method, but is known to achieve good pruning accuracy.
\OURS achieves higher accuracy as compared to NN slimming, and comparable accuracy to DCP. As for the latter, the benefit of \OURS is that we do not need to perform
any greedy channel selection and the entire Hessian calculation and channel selection is performed in one pass.\footnote{We also tried to test DCP on other models
but the code base is old and we were not able to use it for WideResNet32 or PreResNet29. As such we considered other pruning methods, besides EigenDamage, for comparison with those models.}

For PreResNet29, we also compare with NN Slimming~\cite{liu2017learning}, to compare with prior reported results on this model. 
We observe that \OURS achieves up to 6\% higher accuracy as compared to NN Slimming method, and slightly higher accuracy as compared to EigenDamage.
It is interesting to note that \OURS can keep the accuracy the same as baseline, up to
pruning 70\% of the parameters (corresponding to 30\% remaining parameters in~\fref{fig:cifar10_pareto}).

We also present results on VGG16 and compare with other works in the literature, including GAL~\cite{lin2019towards}, HRank~\cite{lin2020hrank}, and VarP~\cite{zhao2019variational},
as reported in Table~\ref{tab:vgg16}.
Here, we consistently achieve higher accuracy.
In particular, \OURS with 29.7\% FLOPs and 10.1\% parameters achieves the highest accuracy (despite 
using a pretrained model with lower baseline accuracy).
Similarly, \OURS with 20.3\% FLOPs and 5.1\% parameters achieves 93.37\% accuracy,
with less than $2\times$ FLOPs and $3\times$ fewer parameters as compared to HRank in the same block.
For extreme pruning, \OURS achieves 91.22\% accuracy with only 1.6\% of the parameters remaining.
To the best of our knowledge, this level of aggressive pruning, while maintaining such high accuracy, has not been reported in the literature.

It is interesting to visualize how \OURS performs channel selection
using the second-order sensitivity discussed in Sec.~\ref{subsec:hap}, and the result can be found in~\secref{sec:appendix_extra_results}

% ------------------------------------------------------------------------
% \vspace{-3mm}
\subsection{Neural Implant Results on CIFAR-10}
\label{sec:channel_condense}
% \vspace{-2mm}

Despite \OURS's competitive results as compared to prior pruning methods,
it still has lower accuracy as compared to baseline. This is known problem and
shortcoming of structured pruning methods.
We propose to use a low rank Neural Implant to address this problem, and find
it particularly helpful for moderate levels of structured pruning.
In particular, for the CNNs tested in this paper we replace sensitive
$3\times3$ spatial convolutions with a pointwise convolution.
This replacement still reduces the number of parameters for the $3\times3$ convolution
by a factor of $9\times$.

We repeated the previous experiments with this approach, and report the results in~\fref{fig:cifar10_pareto} (blue line).
We observe that \NIOURS consistently achieves better performance than \OURS for both the 
same parameter size (first row) and the same FLOPs (second row), which also surpasses the
performance of DCP~\cite{zhuang2018discrimination} that has a competitive result with \OURS.

For some cases, the performance of the pruned network slightly exceeds the baseline accuracy. 
In particular, for ResNet56, we observe up to 1.5\% higher accuracy as compared to
\OURS, and up to 2\% higher accuracy as compared to EigenDamage.
We observe a similar trend for both WideResNet32/PreResNet29, where \NIOURS consistently performs better than both \OURS and EigenDamage. 

It should be noted that the gains from \NIOURS diminish for higher pruning ratios (around
20\% remaining parameters for ResNet56, and around 30\% remaining for WideResNet32/PreResNet29). 
This is expected, since there is a trade-off associated with adding the Neural Implant. 
While the implant helps reduce the information loss from completely removing sensitive 
channels, it does so by adding additional parameters.
As such, we actually have to enforce a larger pruning ratio to meet a target
model size. 
As the channels are sorted based on their sensitivity (from~\eref{eq:trace_approximation}), this means
that we have to prune the next set of more sensitive channels to satisfy the target.
However, if such channels have much higher sensitivity, then that can actually degrade the performance. This is what happens for extreme pruning cases, since most of the remaining parameters will be
highly sensitive; and, as such, the gains achieved by the Neural Implant
will not be enough.

In addition to parameter percentage, we also compare \NIOURS results based on remaining FLOPs 
with other methods reported in the literature.
This is shown in~\tref{tab:dhap}.
As one can see, with a high remaining FLOPs percentage, \NIOURS can reach 93.55\% accuracy with 
only 0.33\% degradation as compared with the corresponding pretrained baseline model.
It should be noted that state-of-the-art methods such as FPGM~\cite{he2019filter} and 
LFPC~\cite{he2020learning} requires 6.4\% more FLOPs to reach comparable performance.
Moreover, when the target percentage of remaining FLOPs is small, \NIOURS only incurs 0.96\% 
accuracy degradation as compared with 2.31\% for \OURS and 2.54\% for HRank~\cite{lin2020hrank} (with comparable FLOPs and baseline accuracy).

% ----------------------------------------------------------------------
% \vspace{-2mm}
\subsection{HAP Results on ImageNet}
\label{subsec:hap_result_imagenet}
% \vspace{-2mm}

We also test \OURS on ImageNet using ResNet50, and report the
results in~\tref{tab:imagenet}.
We compare with several previous structured pruning methods including SSS~\cite{huang2018data}, 
CP~\cite{he2017channel}, ThiNet~\cite{luo2017thinet}, and HRank~\cite{lin2020hrank}.
It should be noted that the accuracy of our pretrained baseline is slightly lower than
HRank, yet our \OURS method still achieves higher accuracy.
For instance, in all cases, \OURS achieves higher accuracy with smaller
number of parameters as compared to all prior work reported on ResNet50.
The highest difference corresponds to 34.74\% remaining parameters (i.e., pruning 65.26\% of parameters), where \OURS has 2\% higher Top-1 accuracy with 19.26\% fewer parameters as compared
to  HRank (although for fairness our FLOPs are slightly larger).
We observe a consistent trend even for high pruning ratios. For example,
with 20.47\% remaining parameters, \OURS still has more than 2\% higher accuracy as compared to HRank.
We should also note that despite using second-order information, \OURS is quite efficient, and the end-to-end Hessian calculations were completed
three minutes on a single RTX-6000 GPU.

\begin{minipage}{\textwidth}

    \begin{minipage}[t]{0.48\textwidth}
    \makeatletter\def\@captype{table}
    \caption{Comparison between \OURS and \NIOURS, and other
    state-of-the-art pruning methods on ImageNet.
    Here, ResNet50 is the baseline used in HRank~\cite{lin2020hrank}'s table, while ResNet50-\OURS is the baseline used by \OURS.}
    \begin{center}
    \begin{adjustbox}{width=1.0\linewidth} 
    \begin{tabular}{lcccccccccccccc} \toprule
    \label{tab:imagenet}

    Method                              & Top-1                 & Param.(\%)  & FLOPs(\%)           \\ 
    \midrule
    \ha  ResNet50-MetaP                 & 76.16                 & 100.0        & 100.0              \\
    \ha  ResNet50                       & 76.15                 & 100.0        & 100.0              \\ 
    \hc  ResNet50-\OURS                 & 75.62                 & 100.0        & 100.0              \\ 
    \midrule
    \ha  SSS-32\cite{huang2018data}     & 74.18                 & 72.94        & 68.95              \\ 
    \ha  CP\cite{he2017channel}         & 72.30                 & -            & 66.75              \\ 
    \ha  GAL-0.5\cite{lin2019towards}   & 71.95                 & 83.14        & 56.97              \\
    \ha  MetaP\cite{liu2019metapruning} & 72.27                 & 61.35         & 56.36             \\
    \ha  HRank\cite{lin2020hrank}       & 74.98                 & 63.33        & \bf{56.23}         \\ 
    \hc  \OURS                          & 75.12                 & 55.41         & 66.18             \\ 
    \hc  \NIOURS                        & \bf{75.36}            & \bf{53.74}   & \bf{55.49}         \\

    \midrule
    \ha  GDP-0.6\cite{lin2018accelerating}      & 71.19            & -             & 45.97     \\
    \ha  GDP-0.5\cite{lin2018accelerating}      & 69.58            & -             & 38.39     \\
    \ha  SSS-26\cite{huang2018data}             & 71.82            & 61.18         & 56.97     \\
    \ha  GAL-1\cite{lin2019towards}             & 69.88            & 57.53         & 38.63     \\
    \ha  GAL-0.5-joint\cite{lin2019towards}     & 71.82            & 75.73         & 44.99     \\
    \ha  HRank\cite{lin2020hrank}               & 71.98            & 54.00         & \bf{37.90}\\
    \hc  \OURS                                  & \bf{74.00}       & \bf{34.74}    & 40.44     \\
    
    \midrule
    \ha  ThiNet-50\cite{luo2017thinet}          & 68.42            & 33.96         & 26.89     \\
    \ha  GAL-1-joint\cite{lin2019towards}       & 69.31            & 40.04         & 27.14     \\
    \ha  HRank\cite{lin2020hrank}               & 69.10            & 32.43         & \bf{23.96}\\
    \ha  MetaP\cite{liu2019metapruning}         & 70.07            & 41.44         & 25.69     \\
    \hc  \OURS                                  & \bf{71.18}       & \bf{20.47}    & 32.85     \\
    \bottomrule
    \end{tabular}
    \end{adjustbox}
    \end{center}
    \end{minipage}\hfill
    \begin{minipage}[t]{0.48\textwidth}
    \makeatletter\def\@captype{table}
    \caption{Ablation study on the sensitivity metric. R-\OURS denotes pruning by reversely using sensitivity in \OURS. Random is conducted by randomly allocating channel-wise sensitivity. 
    }
    \begin{center}
    \begin{adjustbox}{width=1.0\linewidth} 
    \begin{tabular}{lcccccccccccccc} \toprule
    \label{tab:ablation}
    Method & Acc. & Param.(\%) & FLOPs(\%) & Channel    \\
    
    \midrule
    
    R-\OURS    & 89.77      & 47.48         & 46.98         & 65  \\
    Random     & 93.12      & 45.60         & 46.68         & 60  \\
    Magnitude  & 93.29      & 61.82         & 39.29         & 55  \\
    \hc \OURS  & \bf{93.38} & \bf{41.53}    & \bf{38.74}    & 50  \\
    
    \midrule
    
    R-\OURS     & 89.97      & 42.85        & 39.99         & 60  \\
    Random      & 92.21      & 36.01        & 33.99         & 50  \\
    Magnitude   & 92.99      & 56.15        & 34.97         & 50  \\
    \hc \OURS   & \bf{93.23} & \bf{35.50}   & \bf{33.99}    & 45  \\
    
    \midrule
    
    R-\OURS    & 88.83       & \bf{27.15}   & 30.61         & 50    \\
    Random     & 90.95       & 29.64        & 31.32         & 40    \\
    Magnitude  & 92.45       & 47.28        & 28.97         & 42.2  \\
    \hc \OURS  & \bf{92.81}  & 31.08        & \bf{28.85}    & 40    \\
    
    \midrule
    
    R-\OURS     & 88.18       & 23.34       & 28.04         & 45  \\
    Random      & 90.18       & 25.33       & 25.69         & 30  \\
    Magnitude   & 91.65       & 38.25       & 22.93         & 35  \\
    \hc \OURS   & \bf{92.06}  & \bf{22.05}  & \bf{22.86}    & 31  \\

    \bottomrule 
    \end{tabular}
    \end{adjustbox}
    \end{center}
    \end{minipage}\hfill
\end{minipage}

\begin{table}[!htb]
\begin{minipage}[t]{.48\linewidth}
\caption{\small
       Comparison of \OURS and the gradient-based~\cite{michel2019sixteen} heads pruning methods for RoBERTa-base, evaluated on MRPC and QNLI. 
       As an evaluation metric, we report accuracy for QNLI and the average of accuracy and F1 score for MRPC.
    }
\label{tab:roberta_result}
\begin{adjustbox}{width=1.0\linewidth} 
\subtable[MRPC]
    {
    \setlength{\tabcolsep}{2.5pt}{
       \begin{tabular}{lcccccc}
        \toprule
        \ha  {Parameter (\%)} & 100 & 80  &  60  & 50 & 40 \\%& 30     \\ 
        \midrule         
        \ha Gradient-based~\cite{michel2019sixteen}  
                  & 92.06 & \textbf{92.19} & 89.52 & 89.36 & 89.07 \\%& \textbf{88.68} \\
        \hc \OURS & 92.06 & 91.97 & \textbf{90.74} & \textbf{90.43} & \textbf{89.89} \\%& 88.49 \\
        \midrule         
        \ha  Diff  & 0.00 & -0.22 & +1.22 & +1.07 & +0.82 \\%& -0.19 \\
        \bottomrule
        \end{tabular} 
        }
        }
\end{adjustbox}

\begin{adjustbox}{width=1.0\linewidth} 
\subtable[QNLI]
{
    \setlength{\tabcolsep}{2.5pt}
    {
       \begin{tabular}{lcccccc}
        \toprule
        \ha  {Parameter (\%)} & 100 & 80  &  60  & 50 & 40 \\ %& 30     \\ 
        \midrule         
        \ha Gradient-based~\cite{michel2019sixteen}  
                  & 93.12 &  92.4 & 91.78 & 91.71 & 91.38 \\%& 90.33\\
        \hc \OURS & 93.12 & \textbf{93.30} & \textbf{92.93} & \textbf{92.59} & \textbf{92.27} \\%& \textbf{90.85} \\
        \midrule         
        \ha  Diff  & 0.00 & +0.90 & +1.15 & +0.88 & + 0.89 \\%& +0.52\\
        \bottomrule
        \end{tabular} 
        }
}
\end{adjustbox}
\end{minipage}\hfill
\begin{minipage}[t]{.48\linewidth}
\caption{\small
        Hessian Aware Low Rank (Neural Implant) vs Low Rank.
        Ablation study on Hessian aware Neural Implant. The R-\NIOURS uses low rank replacement without using Hessian information. Specially, the Neural Implant is applied on channels that are most sensitive.
    }
    \label{tab:ablation3}
    % \vspace{-1mm}
    \begin{center}
    \begin{adjustbox}{width=1.0\linewidth} 
        \tiny
    \begin{tabular}{lcccccccccccccc} \toprule

    ResNet56/Cifar10              & Acc.          & Param.(\%)    & FLOPs(\%)    \\
    \midrule
    \ha Low Rank        & 90.79         & 39.56         & 40.33         \\
    \hc \NIOURS         & \bf{93.52}    & \bf{34.49}    & \bf{37.15}     \\
    
    \midrule
    \ha Low Rank        & 90.34         & 34.39         & 37.24         \\
    \hc \NIOURS         & \bf{93.40}    & \bf{29.87}    & \bf{33.20}     \\
    
    \midrule
    \ha Low Rank        & 89.83         & 23.05         & 25.16             \\
    \hc \NIOURS         & \bf{92.92}    & \bf{21.92}    & \bf{24.90}        \\

    \bottomrule 
    \end{tabular}
    \end{adjustbox}
    \end{center}
\end{minipage}\hfill
\end{table}

% ----------------------------------------------------------------------
% \vspace{-2mm}
\subsection{HAP Results on RoBERTa}
\label{subsec:hap_result_roberta}

Additionally, we show in this section that our \OURS method can be applied to Transformer~\cite{vaswani2017attention} architectures to prune unnecessary attention heads from multi-head attention layers.
The results are reported as \tref{tab:roberta_result}, where we test over different  heads prune ratio.
As shown in the table, HAP outperforms the gradient-based method with 40$\sim$60\% prune ratio for MRPC and with all prune ratio for QNLI by a noticeable margin of $\sim$1 point.
The results indicate that Hessian could be more informative than gradient in determining sensitivities of attention heads.

% ------------------------------------------------------------------------------------------
% \vspace{-2mm}
\subsection{Ablation Study}
\label{sec:ablation_study}
% \vspace{-2mm}

We conducted several different ablation experiments to study the
effectiveness of the second-order based metric in \OURS.
For all the experiments, we use ResNet56 on CIFAR-10.

%%%%%%%%%%%%%%%%%%%%%%%%%%%%%%%%%%%%%%%%%%%%%%%%%%%%%%%%%%%%%%%%%%%%%%
%%%%%%%%%%%%%%%%%%%%%%%%%%%%%%%%%%%%%%%%%%%%%%%%%%%%%%%%%%%%%%%%%%%%%%

One of the main components of \OURS is the Hessian trace metric used to sort different 
channels to be pruned.
In particular, this ordering sorts the channels from the least sensitive to most sensitive,
computed based on~\eref{eq:trace_approximation}. In the first ablation study,
we use the reverse order of what \OURS recommends, and denote this method
as R-\OURS.
The results are shown in~\tref{tab:ablation}. It can be clearly observed that for all cases
R-\OURS achieves lower accuracy as compared to \OURS (more than 3\% for the
case with 35.50\% remaining parameters).
In the second ablation experiment, we use a random order for pruning the layers, irrespective
of their second-order sensitivity, and denote this 
as Random in~\tref{tab:ablation}. Similar to the previous case, the random
ordering achieves consistently lower accuracy as compared to \OURS.
In addition, its results exhibit a larger variance.

Another important ablation study is to compare the performance of the Hessian-based pruning with the commonly
used magnitude based methods that use variants of $\|w\|^2_2/p$ (denoted as Magnitude in~\tref{tab:ablation}).
To make a fair comparison, we set the FLOPs of the model after pruning
to be the same for \OURS and the magnitude based pruning (and slightly higher for the latter to be fair).
The results are reported in~\tref{tab:ablation}.
As the results show, \OURS achieves the same accuracy as magnitude based pruning but with much fewer
parameters (i.e., higher pruning ratio). In particular, for pruning with 22.53\% of FLOPs (last row of~\tref{tab:ablation}),
\OURS achieves 92.06\% which is almost the same as magnitude based pruning (91.65\%).
However, \OURS achieves this accuracy with only 21.01\% of the parameters remaining, as compared to 38.25\%, which is quite a
significant~difference.
This is expected, as \OURS's performance was higher than the different
magnitude based results reported in the literature, for both
the CIFAR-10 and ImageNet tests of the previous subsections (Sec.~\ref{sec:hap_cifar10_results} and~\ref{subsec:hap_result_imagenet}, respectively).

To study the role played by Hessian analysis in Neural Implant, we've run another set of experiment to prove that Hessian-aware Neural Implant is a combination of sensitivity analysis and Low Rank approximation. The way of using Hessian to guide where Low Rank approximation happens is way more effective than directly applying it to the original model.
The key idea is
to only apply low rank for sensitive layers, as measured by Hessian trace.
We compare Neural Implant with
Low Rank in~\tref{tab:ablation3}, which clearly shows that we can get up to 2\% higher
accuracy with lower Params/FLOPs.

%% file: _s5_conclusion.tex
% \vspace{-4mm}
\section{Conclusion}
\label{sec:conclusion}
% \vspace{-2mm}

% ---------------------------

Existing structured-pruning methods often result in significant 
accuracy degradation for moderate pruning levels. 
To address this, we propose \OURS, a new second-order structured-pruning method that uses
the Hessian trace as the sensitivity metric for pruning a NN model.
We also proposed a new Neural Implant approach that uses \OURS's sensitivity
metric to perform targeted replacement of sensitive neuron's with a light-weight
low rank implant.
The main intuition is to prune insensitive components and to use the
Neural Implant for moderately sensitive components., instead of completely pruning them.
We performed extensive empirical tests using multiple NN models.
We compared with several prior works, including both the second-order based structured pruning method of EigenDamage, as well as several magnitude-based pruning methods. 
\OURS consistently achieved higher accuracy with fewer parameters.
Specifically, \OURS achieves 94.3\% accuracy ($<0.1\%$ degradation) on PreResNet29
(CIFAR-10), with more than 70\% of parameters pruned.
In comparison to EigenDamage, we achieve up to 1.2\% higher accuracy with fewer parameters and FLOPs.
Moreover, for ResNet50 \OURS achieves 75.1\% top-1 accuracy (0.5\% degradation) on 
ImageNet, after pruning almost half of the parameters.
In comparison to the prior state-of-the-art of HRank, we achieve up to 2\% higher accuracy with fewer parameters and FLOPs.
For head-pruning of RoBERTa, \OURS can achieve more than 0.8\% better performance as compared to previous gradient based method with 60\% pruned heads. 
We have open sourced our implementation available at~\cite{HAP}.

%% file: _appendix.tex
\section{Solving~\eref{eq:objective}}
\label{sec:lagrangian}
\eref{eq:objective} can be solved by forming the corresponding Lagrangian
and finding its saddle points:
\begin{equation}
\small
\begin{split}
    &\Loss = \frac{1}{2} \Delta w ^T H \Delta w + \lambda^T (\Delta w_{p} + w_p),\\
    &\frac{\partial \Loss}{\partial \Delta w} = H \Delta w + \begin{pmatrix} \lambda\\0 \end{pmatrix} = 0,\\
    &\begin{pmatrix} H_{p,p} & H_{p,l} \\ H_{l,p} & H_{l,l} \end{pmatrix} 
    \begin{pmatrix} \Delta w_{p}\\ \Delta w_{l} \end{pmatrix} + \begin{pmatrix} \lambda\\0 \end{pmatrix} = 0,
\end{split}
\end{equation}
where $\lambda \in \R^p$ is the Lagrange multiplier.
By expanding this equation, we get:
\begin{align}
\small
    \label{eq:delta_w_l}
    H_{p, p} \Delta w_{p} + H_{p, l} \Delta w_{l} + \lambda &= 0,\\
    \label{eq:delta_w_l_2}
    H_{l, p} \Delta w_{p} + H_{l, l} \Delta w_{l} &= 0.
\end{align}
Using the constraint in~\eref{eq:objective} and adding it to~\eref{eq:delta_w_l_2}, we have:
\begin{equation}
\small
\begin{split}
\label{eq:delta_w_-l2}
    -H_{l, p} w_p + H_{l, l} \Delta w_{l} = 0, \\
    \Delta w_{l} =  H_{l, l}^{-1} H_{l, p} w_p.
\end{split}
\end{equation}
This equation gives us the optimal change to the unpruned parameters ($w_l$), if
a pre-selected set of weights is pruned ($w_p$).
Inserting this into~\eref{eq:objective}, results in the following:
\begin{equation}
\small
    \frac{1}{2} \Delta w ^T H \Delta w = \frac{1}{2} w_{p}^T (H_{p, p} - H_{p, l}  H_{l, l}^{-1} H_{l, p}) w_{p}.
\end{equation}

\section{Detailed Experimental Setup}
\label{sec:detailed_experimental_setup}
Here we present the details of the experiments performed in the paper.

\textbf{Computer Vision} 
For model pretraining on CIFAR-10~\cite{krizhevsky2010cifar}, we use the same setting as EigenDamage~\cite{wang2019eigendamage}. 
To finetune the pruned model for performance improvement, we use SGD with momentum 0.9 and train the compressed model for 160 epochs for CIFAR-10~\cite{krizhevsky2010cifar} and 120 epochs for ImageNet\cite{krizhevsky2012imagenet}. The initial learning rate is set as 2e-2 for CIFAR-10~\cite{krizhevsky2010cifar}, 1e-3 for ImageNet, and reduce by one-tenth twice at half and 3/4 of the full epoch.
For CIFAR-10~\cite{krizhevsky2010cifar}, we use a batch size of 64 and weight decay of 4e-4, and for ImageNet we use a batch size of 128 and weight decay of 1e-4. We also set a pruning ratio limit for each layer, following~\cite{wang2019eigendamage}.

As for Neural Implant, we select a fixed neural implant ratio of 0.2, meaning that 20\% of the pruned 3x3 convolution kernels are replaced by 1x1 convolution kernels.

We have open sourced our implementation available at~\cite{HAP}.

\textbf{Natural Language Understanding}
In order to compute the Hessian sensitivity of attention heads, we assign all weight matrices in a single head (i.e., query, key, value, and output matrices) into a group of parameters.
Although we prune the least sensitive heads \textit{globally} across all layer, 
we retain at least on head in each layer as we empirically find that removing all heads from a single layer can result in a large accuracy drop.
We evaluate our method on MRPC~\cite{dolan2005automatically} and QNLI~\cite{rajpurkar2016squad} of the GLUE tasks~\cite{wang2018glue}.
We compare our method to the gradient-based heads pruning method of~\cite{michel2019sixteen}.
For both methods, we first finetune the pretained model on downstream tasks until it achieves the best accuracy, apply heads pruning, and perform additional finetuning of 10 epochs to recover accuracy degradation.
We follow the same learning rate and optimizer in RoBERTa~\cite{liu2019roberta} for all the experiments.

\section{Sensitivity Convergence Results}
\label{sec:hutchinson_convergence_speed}

\begin{figure*}[t]
\centering
\includegraphics[width=.3\textwidth]{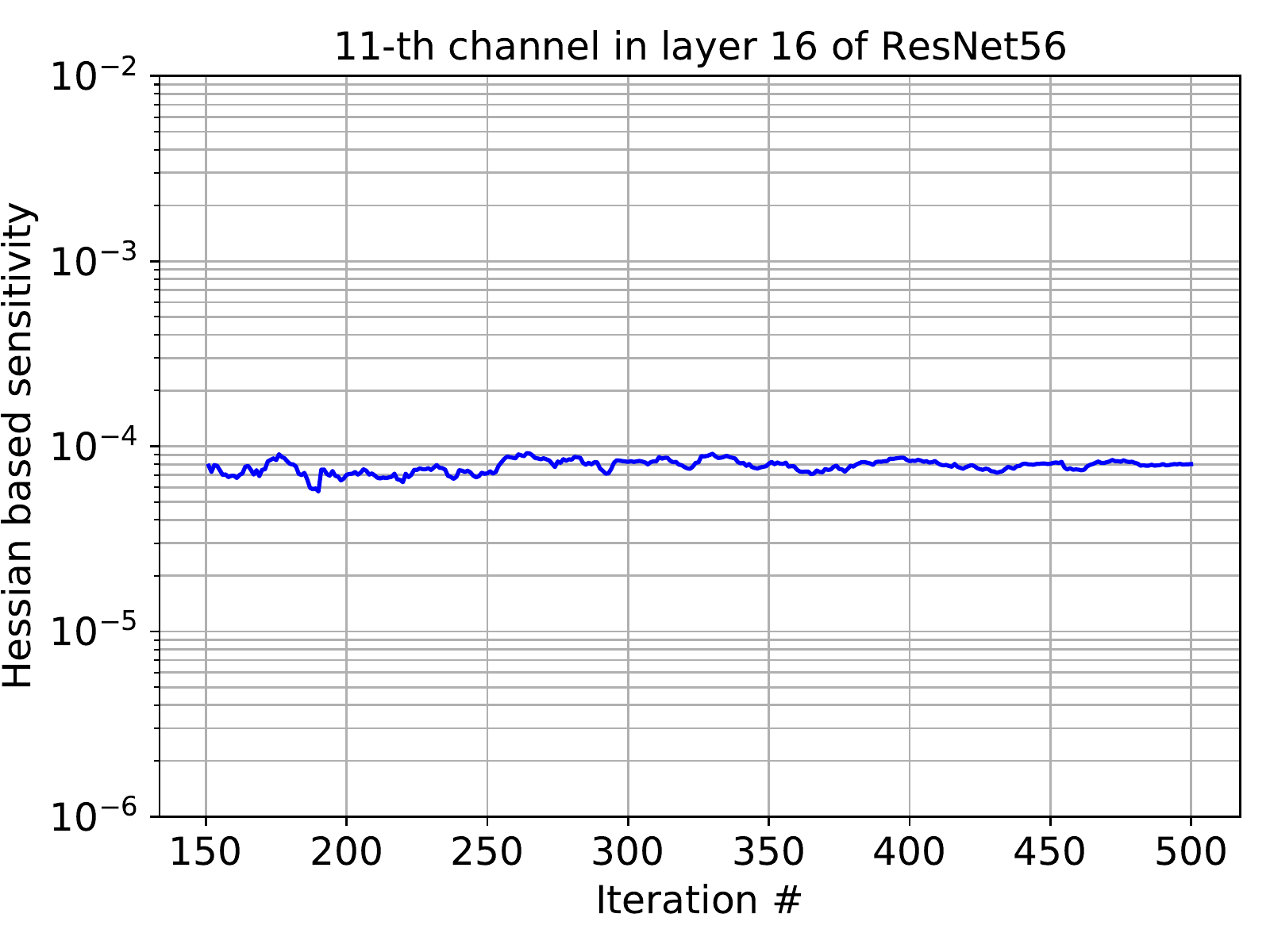}
\includegraphics[width=.3\textwidth]{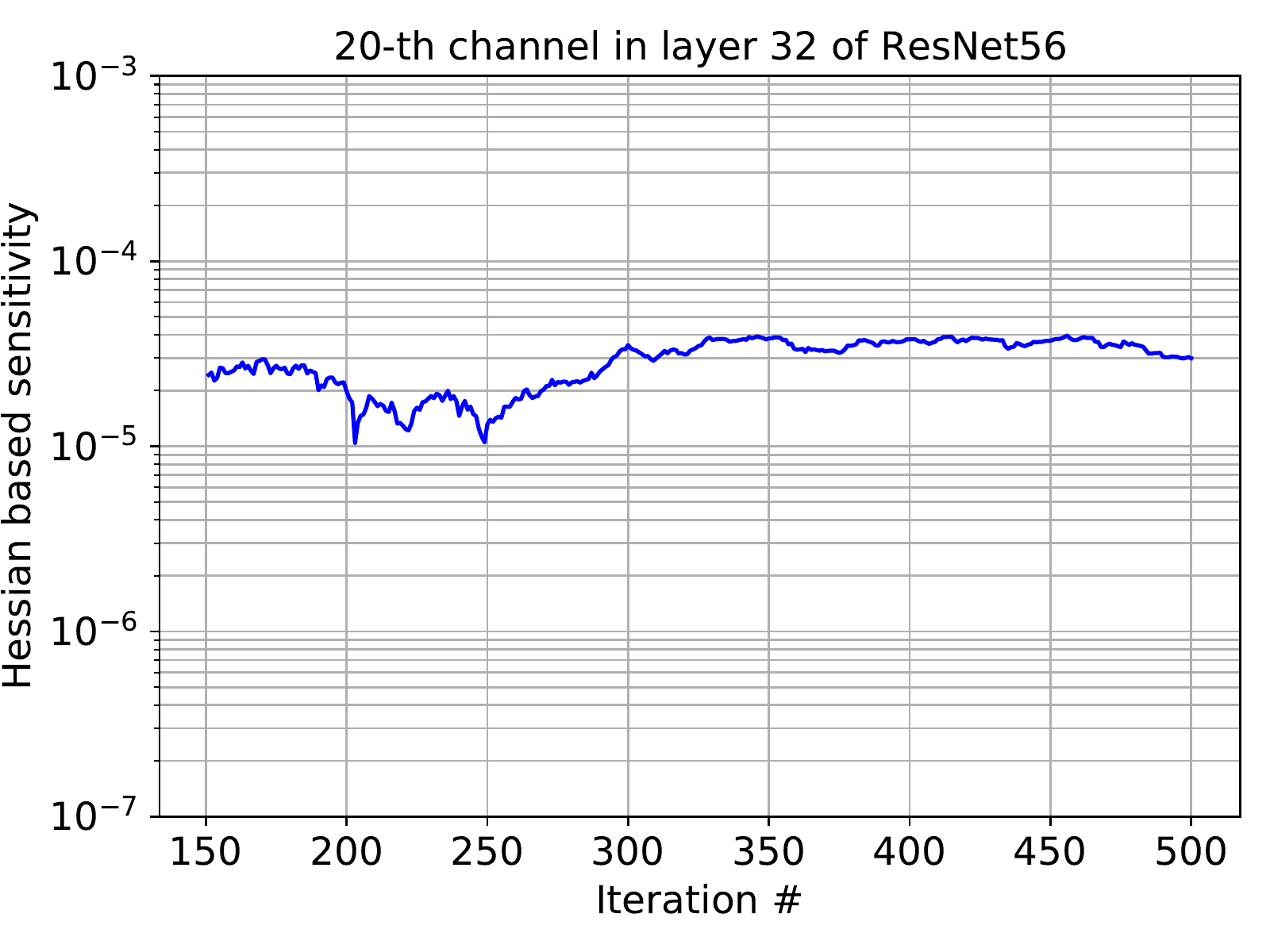}
\includegraphics[width=.3\textwidth]{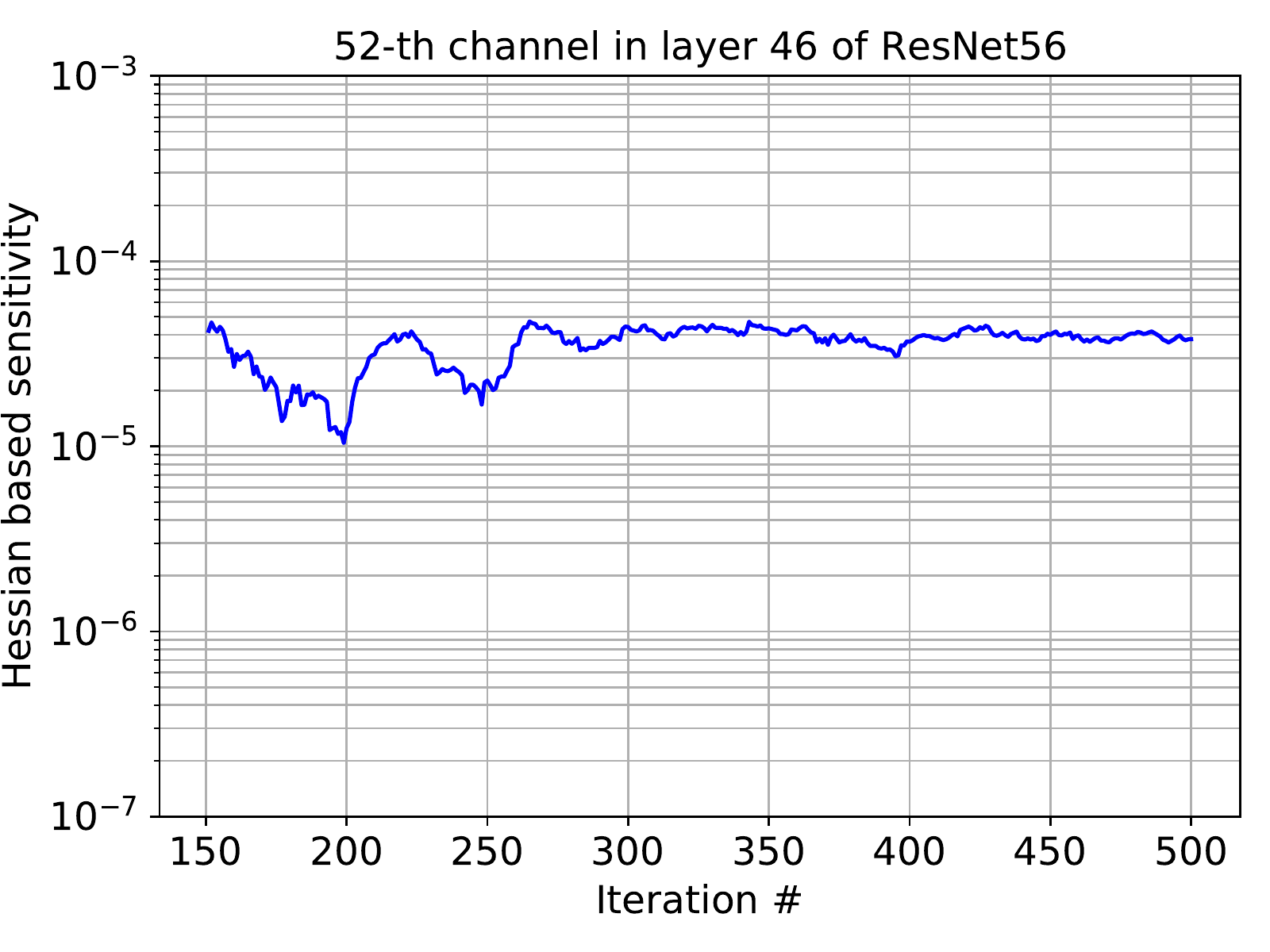}
\caption{
The convergence of Hessian-based sensitivity throughout the Hutchinson iterations, for different channels of ResNet56.
Here, the x-axis is the Hutchinson iteration, and the y-axis is the approximation for the sensitivity corresponding to~\eref{eq:trace_approximation}.
As one can see, the approximation converges after about 300 iterations.
}
\label{fig:resnet56_sensitivity_illustration}
\end{figure*}

As discussed in Section~\ref{sec:method}, we compute the sensitivity based on the trace of the Hessian as presented
in~\eref{eq:trace_approximation}. This approximation can be computed
without explicitly forming the Hessian operator, by using the Hutchinson method~\cite{avron2011randomized,bai1996some}.
In this approach, the application of the Hessian to a random vector is calculated through backpropogation (similar 
to how gradient is backpropagated)~\cite{yao2019pyhessian}.  
In particular, for a given random vector  $v\in R^{p}$ with i.i.d. components, we can show:
\begin{equation}
\small 
    Tr(H) = \E[v^THv].
\end{equation}
See~\cite{yao2019pyhessian,yao2020adahessian} for details and discussion.
We can directly use this identity to compute the sensitivity in~\eref{eq:trace_approximation}:
\begin{equation}
\small 
    \frac{Trace(H_{p,p})}{2p}\|w_p\|_2^2 = \frac{1}{2p}\|w_p\|_2^2\E[v^THv].
\end{equation}
Here, note that the norm of the parameters is a constant. 
One can prove that for a PSD operator, this expectation converges to the actual trace. 
To illustrate this empirically, we have plotted the convergence for this sensitivity metric for different channels of ResNet56.
See~\fref{fig:resnet56_sensitivity_illustration}. 
As one can see, after roughly 300 iterations we get a very good~approximation.

In practice, we set Hutchinson iteration to 300 conform to the result above and 
we show the average time to calculate the channel-wise sensitivity on one Titan-Xp GPU for HAP in~\tref{tab:runtime}. Our method is fast and effective and runs around 100 seconds.

%%%%%%%%%%%%%%%%%%%%%%%%%%%%%%%%%%%%%%%%%%%%%%%%%%%%%%%%%%%%%%%%%%%%%%%%%%%%%%%%%%%
%%%%%%%%%%%%%%%%%%%%%%%%%%%%%%%%%%%%%%%%%%%%%%%%%%%%%%%%%%%%%%%%%%%%%%%%%%%%%%%%%%%

\section{Limitations and Future Work}
\label{sec:limitations}
We believe it is critical for every work to clearly state its limitations, especially in this area. 
An important limitation is that computing the second order information adds some computational overhead. However, the overhead is actually much lower than expected as shown in~\tref{tab:runtime}.
Another limitation is that in this work we only focused on computer vision (image classification) and natural language understanding, but it would
be interesting to see how \OURS would perform for more complex tasks such as object detection and machine translation. 
Third, in this paper, we solely work on static pruning (i.e., the model is fixed after pruning). 
However, for different inputs, the model can be adaptively/dynamically pruned so that we can minimize the accuracy degradation. 
We leave this as a future work.

% -----------------------------------
\section{Additional Results}
\label{sec:appendix_extra_results}
\begin{figure*}[!htbp]
\centering
\includegraphics[width=0.87\textwidth]{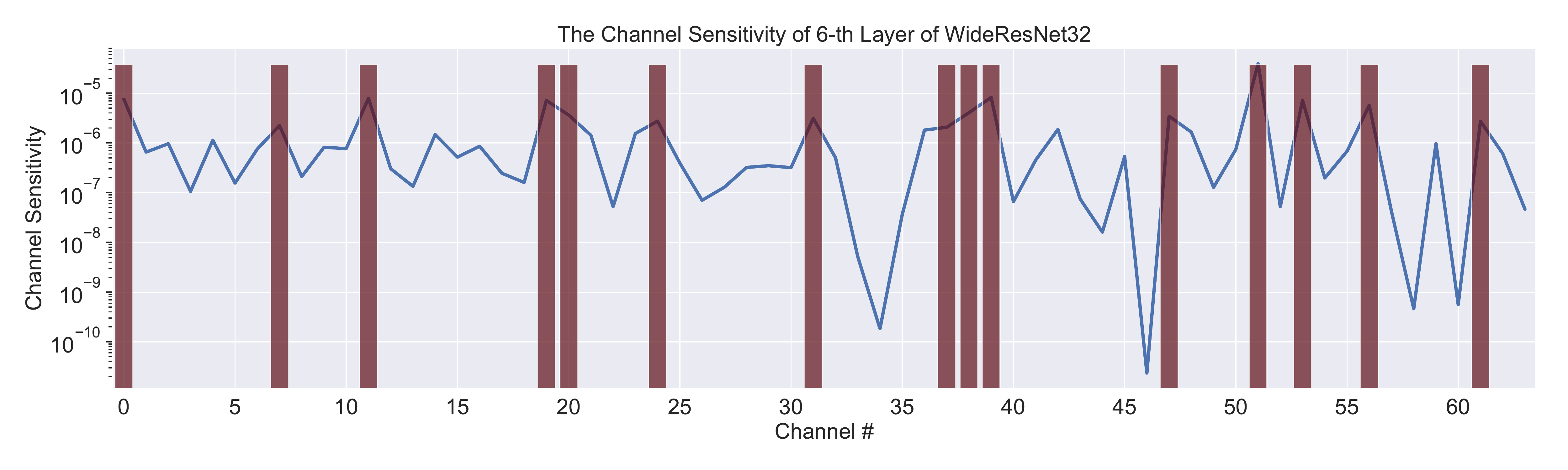}
\includegraphics[width=0.96\textwidth]{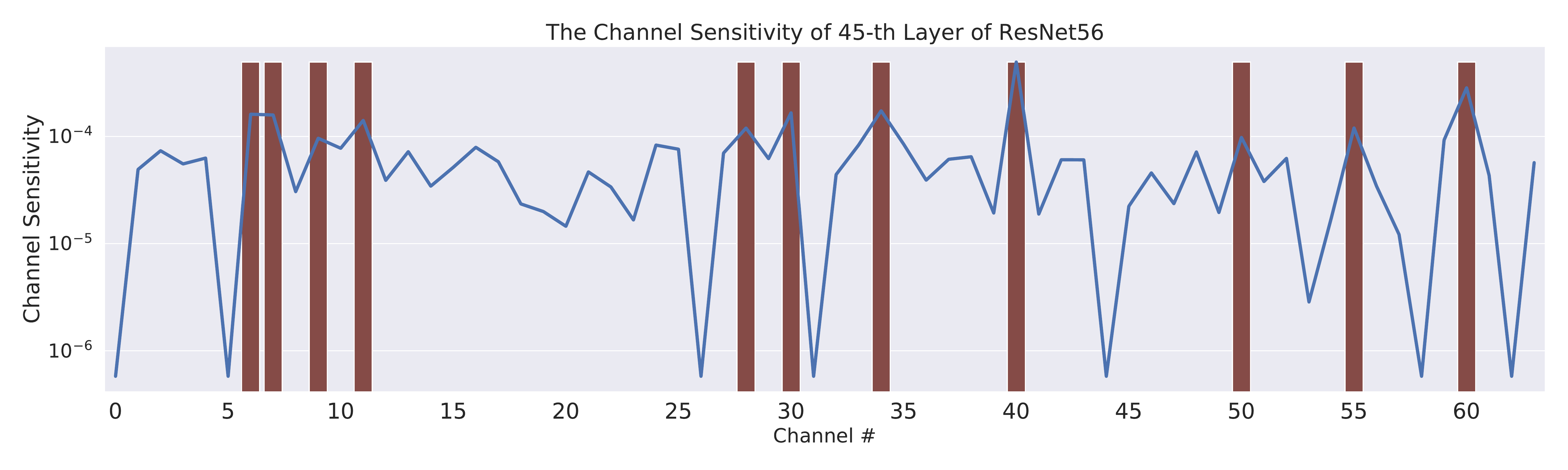}
\includegraphics[width=0.96\textwidth]{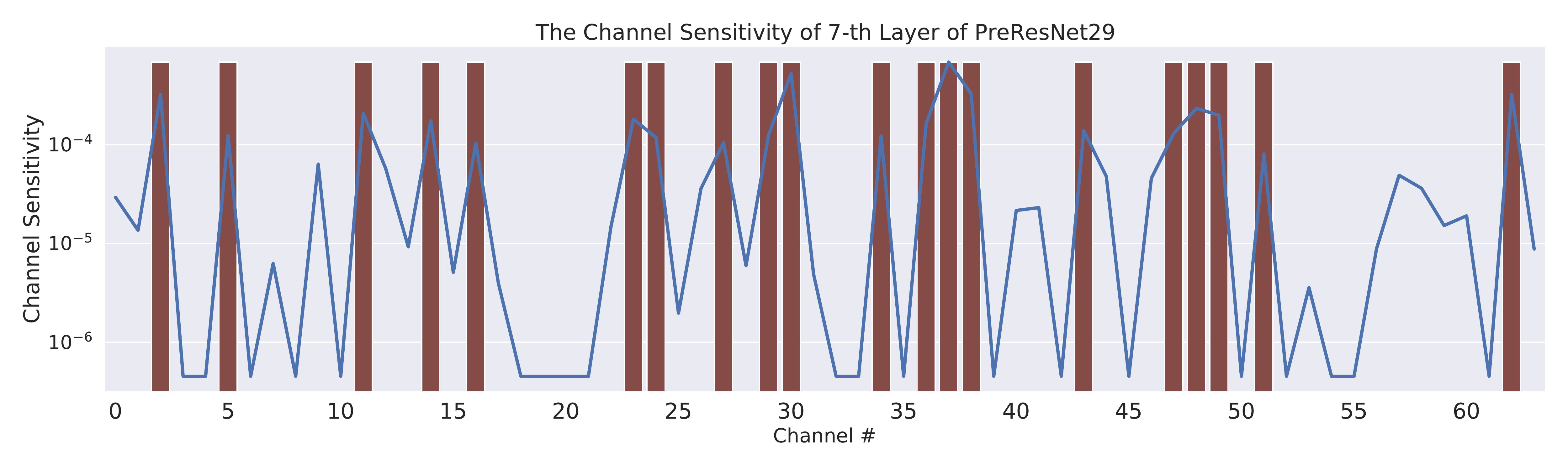}
\caption{
Illustration of sensitivity of the t6h layer of WideResNet32, 45th convolution layer of ResNet56, and the 7th convolution layer of PreResNet29. 
The x-axis denotes the channel index, and the blue line denotes the corresponding second-order sensitivity computed using~\eref{eq:trace_approximation}.
The red bar is added to channels that remain unpruned with the \OURS method. As one can see, these correspond to sensitive channels that have large values on the blue line.
}
  \label{fig:resnet56_PreResNet_sensitivity}
\end{figure*}

Here, we show the distribution of pruning for different channels of WideResNet32, ResNet56, and PreResNet29.
See~\fref{fig:resnet56_PreResNet_sensitivity}.
As one can see, \OURS only prunes insensitive channels, and keeps channels with high sensitivity (computed based on~\eref{eq:trace_approximation}).

%%%%%%%%%%%%%%%%%%%%%%%%%%%%%%%%%%%%%%%%%%%%%%%%%%%%%%%%%%%%%%%%%%%%%%%%%%%%%%%%%%%
%%%%%%%%%%%%%%%%%%%%%%%%%%%%%%%%%%%%%%%%%%%%%%%%%%%%%%%%%%%%%%%%%%%%%%%%%%%%%%%%%%%

\begin{table}
    \caption{
        Calculation time for channel wise sensitivity in \OURS. 
    }
    \label{tab:runtime}
    \begin{center}
    \begin{tabular}{lcccccccccccccc} \toprule
    Method & WideResNet32    & ResNet56  &  PreResNet \\
    
    \midrule
    \hc  HAP    & 120s & 61s  & 81s             \\ 
    
         \bottomrule 
    \end{tabular}
    \end{center}
\end{table}